\documentclass[11pt]{article}

\usepackage{acl}

\usepackage{times}
\usepackage{latexsym}
\usepackage{xspace}
\usepackage{booktabs}  
\usepackage{multirow}  
\usepackage{graphicx}  
\usepackage{subcaption} 
\usepackage{array}     
\usepackage{algorithm}
\usepackage{algpseudocode}
\usepackage{amsmath}
\usepackage{amssymb}
\usepackage{xcolor}

\usepackage[T1]{fontenc}

\usepackage[utf8]{inputenc}

\usepackage{microtype}

\usepackage{inconsolata}
\usepackage[utf8]{inputenc} 

\usepackage{CJKutf8} 
\title{LycheeCluster: Efficient Long-Context Inference with Structure-Aware Chunking and Hierarchical KV Indexing}

\author{
    Dongfang Li\textsuperscript{1}, 
    Zixuan Liu\textsuperscript{2}, 
    Gang Lin\textsuperscript{1}, 
    Baotian Hu\textsuperscript{1}, 
    \and Min Zhang\textsuperscript{1} \\
    \textsuperscript{1}Harbin Institute of Technology \\
    \textsuperscript{2}University of Electronic Science and Technology of China \\
    \texttt{lidongfang@hit.edu.cn}, \texttt{liuzx@std.uestc.edu.cn}
}

\begin{document}
\newcommand{\model}[0]{LycheeCluster\xspace}
\newcommand{\modelfull}[0]{LycheeCluster\xspace}

\maketitle
\begin{abstract}
    The quadratic complexity of the attention mechanism and the substantial memory footprint of the Key-Value (KV) cache present severe computational and memory challenges for Large Language Models (LLMs) processing long contexts. Existing retrieval-based methods often compromise semantic integrity through fixed-size chunking and suffer from inefficient linear scanning. In this paper, we propose \texttt{\model}, a novel method for efficient KV cache management. \texttt{\model} preserves local semantic coherence via boundary-aware chunking and constructs a recursive hierarchical index rooted in the triangle inequality. This design transforms cache retrieval from a linear scan into a theoretically bounded, logarithmic-time pruning process, while a lazy update strategy supports efficient streaming generation. Experiments demonstrate that \model achieves up to a 3.6$\times$ end-to-end inference speedup with negligible degradation in model performance, outperforming state-of-the-art KV cache management methods (e.g., Quest, ClusterKV).~\footnote{ We will release our code and kernels after publication.}
\end{abstract}

\section{Introduction}
    \label{sec:intro}

The context window of Large Language Models (LLMs) has expanded dramatically, evolving from 4K  to over 2M tokens~\cite{qwen3_technical_report,comanici2025gemini,claude-3.7}, enabling transformative applications in long-document understanding and complex reasoning. However, utilizing this massive context introduces a critical latency bottleneck. During the autoregressive decoding phase, the attention mechanism requires scanning  the entire Key-Value (KV) cache history for every generated token~\cite{quest,tidaldecode,tang2024razorattention}. As the sequence length grows, the overhead of loading these massive KV tensors from memory consumes substantial bandwidth, causing the decoding speed to deteriorate significantly despite the availability of  powerful GPUs~\cite{liu2025comprehensive}.
\begin{figure}[t]
    \centering
    \includegraphics[width=\linewidth]{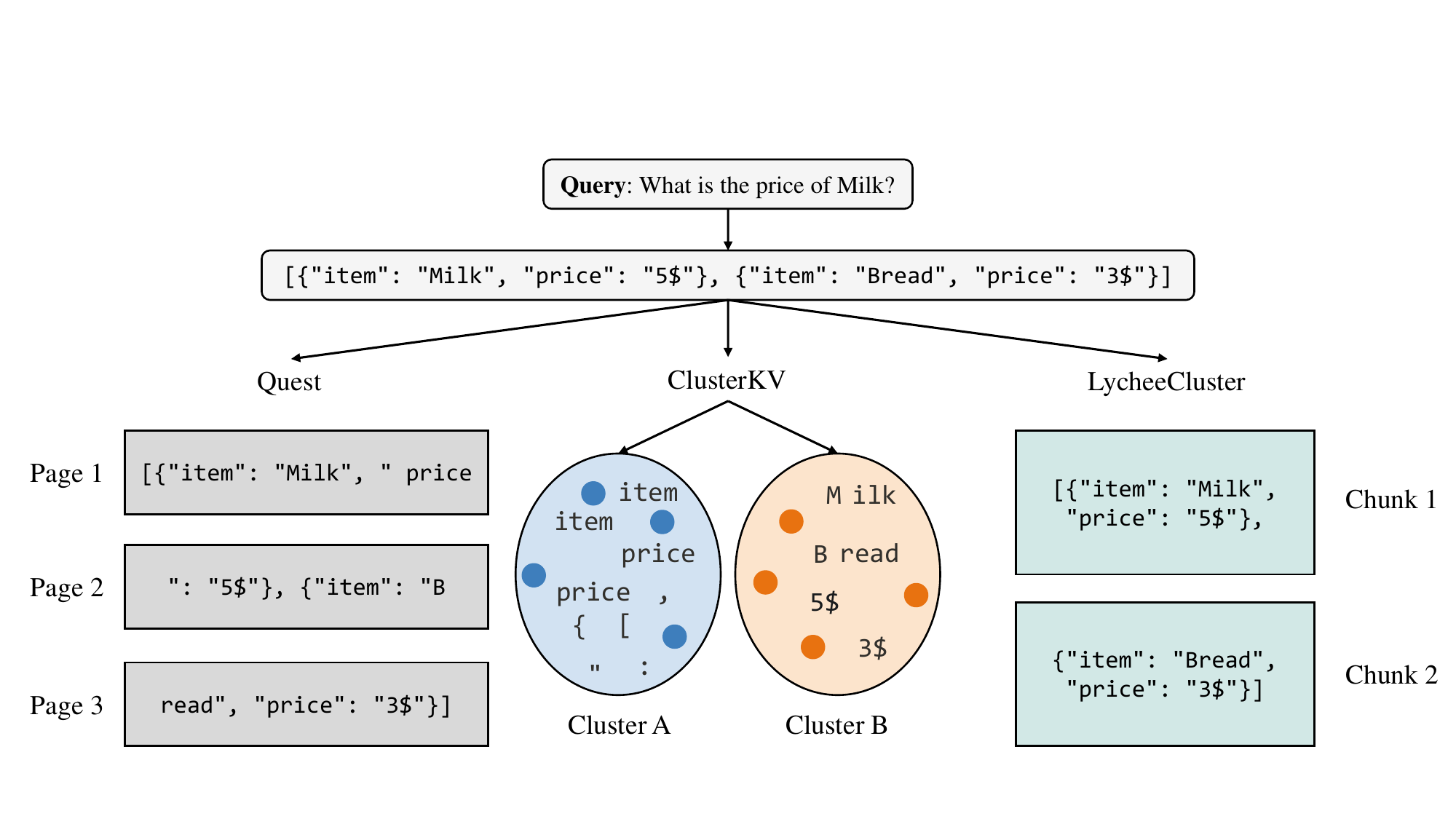} 
    \caption{\textbf{Impact of retrieval granularity on semantic integrity.} We illustrate the limitations of existing methods using a JSON retrieval query. \textbf{(Left) Quest} employs fixed-size pages that arbitrarily sever semantic boundaries. \textbf{(Middle) ClusterKV} uses token-level clustering that scatters locally coherent tokens into disjoint clusters based on vector variance. \textbf{(Right) \model} (Ours) preserves the complete semantic unit via structure-aware chunking, ensuring precise retrieval.}
    \vspace{-4mm}
    \label{fig:concept}
\end{figure}

To address this memory bandwidth bottleneck without permanently discarding information, \textit{Retrieval-based} methods have emerged as a promising direction~\cite{li2024survey,liu2024retrievalattention}. Unlike \textit{Eviction-based} methods (e.g., H2O~\cite{h2o}, StreamingLLM~\cite{streamingllm}) that permanently delete tokens, which often leads to irreversible information loss, retrieval-based approaches preserve the full history and dynamically fetch only a small subset of relevant KV pairs for the attention computation. This paradigm aims to accelerate decoding by transforming the linear-complexity attention mechanism into a sparse operation bounded by a fixed token budget.

However, existing retrieval methods struggle with the granularity of their selection strategies, as conceptually illustrated in Figure \ref{fig:concept}. \textit{Page-based} methods like Quest~\cite{quest} manage KV pairs in fixed-size blocks  (e.g., 64 tokens). This rigid segmentation leads to internal fragmentation, where an entire page is retrieved for a single relevant token, wasting the limited retrieval budget. Conversely, \textit{token-level clustering} methods like ClusterKV~\cite{clusterkv} aggregate tokens based on global semantic similarity. While this reduces fragmentation, treating tokens as isolated vectors disrupts the local structural integrity of the text. For instance, semantically coupled sequences such as code blocks or reasoning steps are scattered across different clusters, preventing the attention head from retrieving the contiguous context necessary for precise reasoning. Furthermore, relying on global clustering algorithms incurs a high update overhead, often forcing the system to rely on stale indices during streaming generation.

In contrast, we hypothesize that the optimal unit for efficient retrieval is neither the arbitrary fixed-size page nor the isolated token, but the \textit{semantically coherent chunk}. To validate this, we conducted a pilot study (\S\ref{sec:motivation}) on the StrucText-Eval~\cite{gu2025structext}. By simply using boundary-aware chunks instead of fixed-size pages, we observed a substantial accuracy improvement (e.g., +15.0\%). This finding suggests that preserving the semantic integrity of retrieval units is the key to maximizing the utility of the sparse attention budget. 

Inspired by this, we propose \texttt{\model}, a novel retrieval-based KV cache management method designed to accelerate long-context decoding through structure-aware indexing. Unlike previous works, \texttt{\model} actively segments the context into variable-length chunks based on natural semantic boundaries (e.g., punctuation, line breaks). To enable sub-linear retrieval complexity, we organize these chunks into the \textbf{hierarchical indices} ($\text{coarse units} \to \text{fine clusters} \to \text{chunks}$), allowing the model to rapidly prune irrelevant branches using a mathematical upper bound of attention scores. Furthermore, to adapt to the streaming nature of decoding, we further introduce a lazy incremental update strategy. During decoding, new chunks are dynamically grafted onto the nearest existing clusters, maintaining index freshness with negligible runtime overhead.

Extensive experiments on long-context understanding and complex reasoning tasks  demonstrate that \texttt{\model} effectively breaks the latency bottleneck without compromising the capability. Specifically, our method achieves up to $3.6\times$ decoding speedup compared to full attention, significantly reducing inference latency for long sequences. Crucially, this efficiency does not come at the cost of precision; \texttt{\model} maintains performance comparable to full attention even under aggressive compression budgets. \texttt{\model} exhibits exceptional robustness in \textsc{RULER}~\cite{hsieh2024ruler} and structured data tasks of \textsc{LongBench V2}~\cite{bai2025longbench}, outperforming token-clustering baselines (e.g., ClusterKV) that often disrupt semantic boundaries. Furthermore, in reasoning-intensive scenarios of \textsc{MATH500}~\cite{lightman2023let}, \texttt{\model} successfully supports the dynamic retrieval needs of chain-of-thought generation, preserving logical coherence where heuristic-based baselines often fall.

Our contributions are summarized as follows:
\begin{itemize}
    \item We identify that structural disruption is a key limitation in existing sparse attention methods and validate structure-aware chunking as a critical solution.
    \item We propose \model, a method featuring hierarchical KV indexing and lazy updates, designed to accelerate inference without compromising semantic integrity.
    \item We achieve state-of-the-art performance with lower latency among retrieval-based methods, offering a robust solution for deploying long-context and reasoning-intensive LLMs.
\end{itemize}

\section{Related Work}
    \paragraph{Eviction-based Sparse Attention.} To address the memory bottleneck of the KV cache, eviction-based methods reduce memory footprint by permanently discarding tokens based on accumulation metrics or positional heuristics~\cite{h2o,streamingllm,snapkv}. While these methods lower latency, the irreversible deletion of tokens leads to information loss, degrading performance in long-context tasks such as multi-step complex reasoning that require full historical recall~\cite{liscbench,kvzip}.

\paragraph{Retrieval-based Sparse Attention.} Retrieval-based approaches preserve the complete KV history and dynamically fetch only relevant subsets for computation~\cite{liu2024retrievalattention,tang2024razorattention,sun2024shadowkv}. For example, Quest~\cite{quest} employs a page-based retrieval strategy, utilizing min-max key statistics of fixed-size blocks to estimate importance. ClusterKV~\cite{clusterkv} adopts a global perspective, grouping tokens into clusters based on vector similarity to accelerate retrieval. 
RazorAttention~\cite{tang2024razorattention} further compresses KV cache by selectively caching tokens in different attention heads.

\paragraph{Semantic Granularity and Integrity.} A critical limitation in existing sparse attention mechanisms is the disruption of semantic units~\cite{fountas2025human,zhu2025sentencekv}. Fixed-size paging arbitrarily severs syntactic boundaries, while token-level clustering fragments locally coherent sequences, such as reasoning steps, into disjoint groups~\cite{chen2025sablock}. Recently, SentenceKV~\cite{zhu2025sentencekv} proposed utilizing natural sentences as the atomic unit for caching. However, their rigid punctuation-based segmentation struggles with structured data (e.g., code, JSON) lacking clear sentence delimiters and suffers from length variance. In contrast, we employ a robust structure-aware chunking strategy that respects semantic boundaries while enforcing size constraints, ensuring stable retrieval efficiency across diverse domains. 
\begin{figure}[t]
    \centering
    \includegraphics[width=0.95\linewidth]{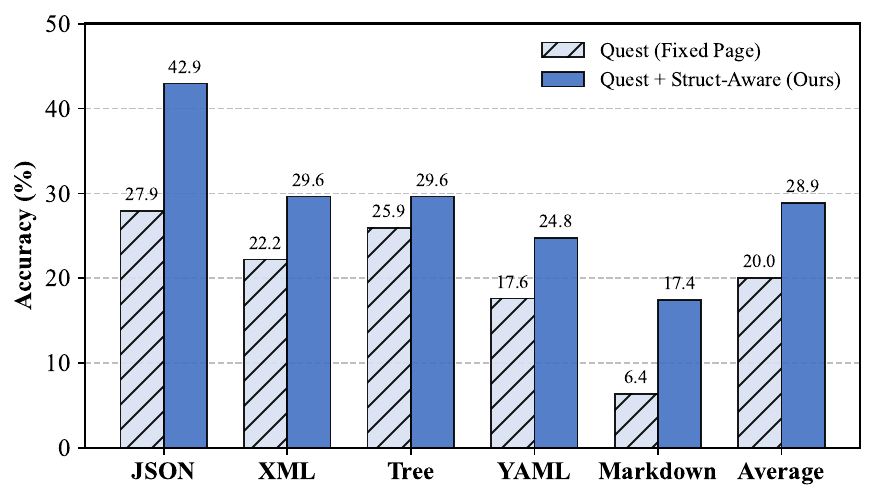}
    \caption{\textbf{Pilot Study on StrucText-Eval.} We compare the standard Quest (fixed page) with a modified version using structure-aware chunks while keeping the scoring metric identical. The significant accuracy gain (e.g., \textbf{+15.0\%} on JSON) confirms that preserving semantic integrity is a prerequisite for effective retrieval.}
    \label{fig:pilot_study}
\end{figure}

\begin{figure*}[h!]
\centering
\includegraphics[width=0.95\linewidth]{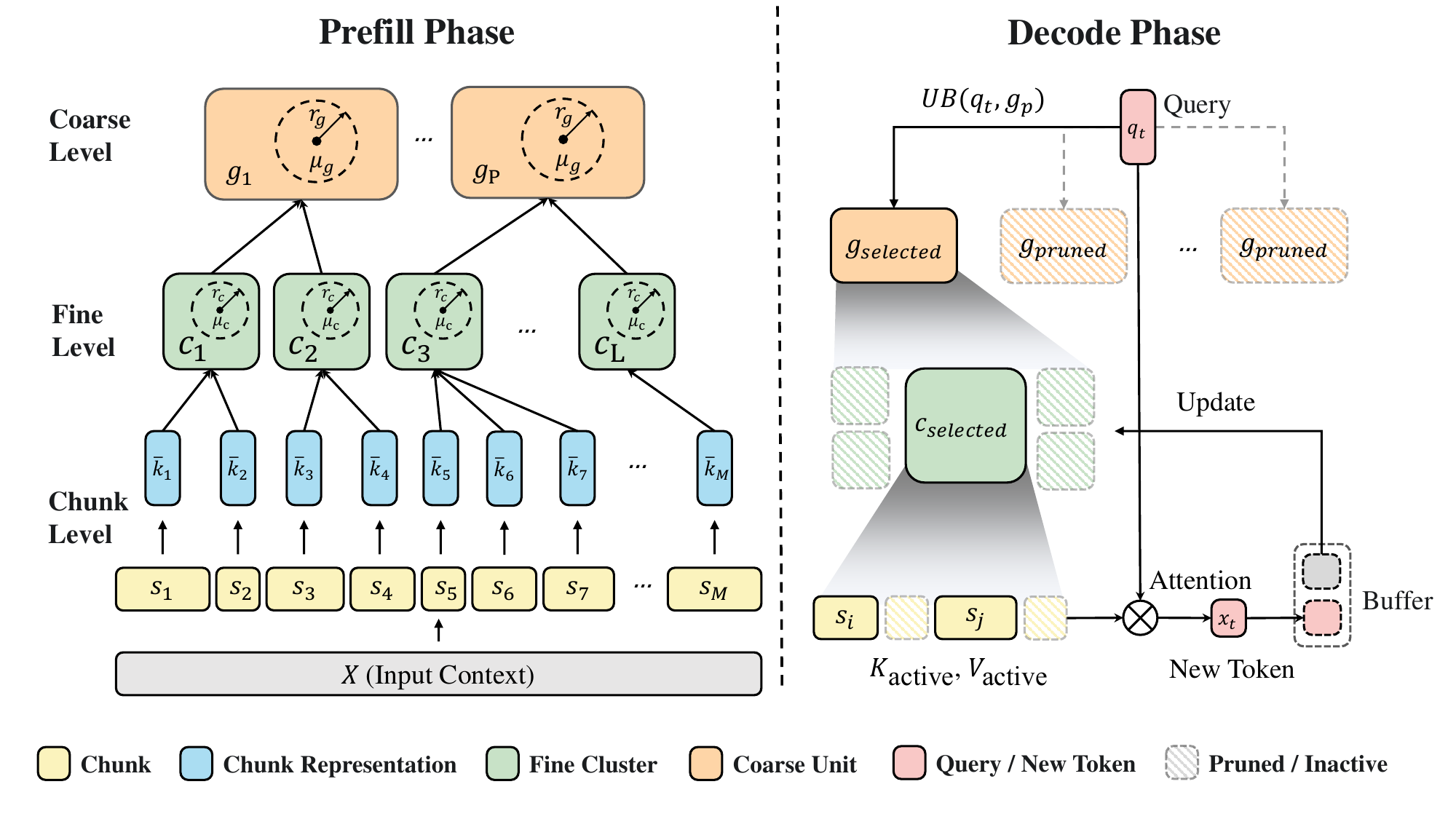}
\caption{The overall pipeline of \texttt{\model}. The left panel illustrates the bottom-up index construction during the prefill phase, where variable-length chunks are hierarchically clustered. The right panel demonstrates the top-down retrieval and incremental update during the decoding phase.}
  \vspace{-0.5cm}
   \label{fig:framework}
\end{figure*}

\section{Motivation}
    \label{sec:motivation}

Current research on sparse attention primarily focuses on optimizing \textit{importance estimation}, i.e., developing better metrics to decide \textit{which} tokens to keep, such as attention accumulation or min-max statistics. However, a fundamental question remains largely unexplored: \textbf{what is the optimal granularity for the atomic unit of scoring?}

Most existing methods adopt rigid segmentation strategies. Quest~\cite{quest} uses fixed-size pages for hardware efficiency, effectively treating the context as a contiguous tape cut at arbitrary intervals. Conversely, ClusterKV~\cite{clusterkv} operates on individual tokens grouped by vector similarity, treating the context as a ``bag-of-words''. As conceptually illustrated in Figure~\ref{fig:concept}, we hypothesize that both approaches compromise semantic integrity: fixed pages sever context at physical boundaries, while token clustering shatters local coherence.

\subsection{A Pilot Study on StrucText-Eval}

To validate the hypothesis that granularity matters as much as scoring, we conduct a controlled pilot study using the StrucText-Eval benchmark~\cite{gu2025structext}. This dataset evaluates LLMs on structured data (e.g., JSON parsing, code completion) where preserving syntactic boundaries is crucial for correctness. We utilize Quest as our experimental baseline due to its representative page-based retrieval mechanism. To isolate the impact of granularity, we kept Quest's importance estimation algorithm (min-max key approximation) unchanged and modified only the segmentation strategy:

\begin{itemize}
    \item \textbf{Baseline (fixed granularity):} standard implementation using fixed pages of size $S=16$.
    \item \textbf{Variant (structure-aware granularity):} we replace fixed pages with variable-length chunks defined by natural delimiters (e.g., line breaks, see Table~\ref{tab:separators}). The average chunk size matched baseline to ensure a fair comparison.
\end{itemize}

\paragraph{Observations.}

The results are summarized in Figure~\ref{fig:pilot_study}. Notably, merely aligning the retrieval units with semantic boundaries yielded a substantial performance boost. On the StrucText-Eval suite, the structure-aware variant outperformed the original Quest by 10.6\% in average accuracy, with gains as high as 15.0\% on strictly formatted tasks like JSON extraction.

\subsection{Implication}

This finding highlights a critical insight: the bottleneck in long-context retrieval is often not the failure to \textit{identify important regions}, but \textit{the fragmentation of those regions}.
When a fixed page cuts a logical unit (e.g., a function definition) in half, or when clustering scatters a KV cache pair, the attention head fails to attend to the complete information even if the retrieval score is high. This \textbf{semantic misalignment} problem renders the retrieved KV cache pairs partially useless.

Therefore, simply patching existing methods is insufficient. We need a retrieval framework designed \textit{natively} around the \textbf{structure-aware chunk} as the atomic unit. This motivation drives the design of \textbf{\model}, which we formally introduce in the next section. By enforcing semantic atomicity at the indexing level, we ensure that every retrieved unit provides a complete, actionable context for the attention mechanism.

\section{Method} 
    In this section, we describe \texttt{\model} in detail, a retrieval-based approach designed to accelerate long-context LLM inference.

\subsection{Formulation}
\label{subsec:formalization}

Given a Transformer model $\mathcal{M}$ and an input sequence $X = \{x_1, \dots, x_{t-1}\}$, the model generates the next token at step $t$ via the attention mechanism. Let $q_t \in \mathbb{R}^d$ be the query vector; the output is a weighted sum of historical KV pairs:
\begin{equation}
    \text{Attn}(q_t, K_{<t}, V_{<t}) = \sum_{i=1}^{t-1} \frac{\exp(q_t^\top k_i / \sqrt{d})}{Z} v_i
\end{equation}
where $Z$ is the normalization factor. Naively computing $q_t^\top k_i$ for all historical tokens leads to linear complexity $\mathcal{O}(t)$. 
Our objective is to approximate this process by retrieving only the most relevant subset of KV pairs. 

As a step towards this, we construct a recursive hierarchical abstraction. Rather than index individual tokens, we organize the KV cache into a tree structure $\mathcal{T}$.
First, we partition the token sequence into a set of semantically coherent chunks $\mathcal{S} = \{s_1, \dots, s_M\}$. For each chunk $s_j$, we define its representative key $\bar{k}_j$ by performing mean pooling over its constituent token keys and projecting the result onto the unit sphere ($\|\bar{k}_j\|_2=1$). The premise is that the strong local coherence of semantic information in   sequences, the relevance of a chunk can be effectively approximated by the relevance of its aggregated representation~\cite{lu2025moba}.
Second, to facilitate efficient search, we build a multi-level index on top of $\mathcal{S}$. We define a parent node $u$ (e.g., a cluster centroid) that covers a set of child nodes $\mathcal{V}_u$ (e.g., chunks or sub-clusters). For each node $u$, we maintain its centroid $\mu_u$ and a covering radius $r_u = \max_{v \in \mathcal{V}_u} \|v - \mu_u\|_2$. Third, leveraging the triangle inequality and the Cauchy-Schwarz inequality, we derive a strict upper bound for the relevance score. For any query $q_t$ and any child node $v$ within the parent node $u$, the dot product similarity satisfies:
\begin{equation}
\begin{aligned}
    q_t^\top v &= q_t^\top (\mu_u + (v - \mu_u)) \\
    &\le \underbrace{q_t^\top \mu_u + \|q_t\|_2 \cdot r_u.}_{\text{Score Upper Bound (UB)}}
\end{aligned}
\label{eq:upper_bound}
\end{equation}

Hence, Eqn.~\ref{eq:upper_bound} serves as the theoretical foundation for our pruning strategy: the similarity between the query and a centroid, plus a slack term determined by the radius, strictly bounds the maximum possible similarity for any element within that cluster. This allows us to safely discard entire branches of the index tree if their upper bound is sufficiently low, without inspecting individual chunks or tokens.
\begin{algorithm}[t]
\caption{Inference Process with \model}
\label{alg:inference}
\small
\textbf{Input:} Prompt $X$, Model $\mathcal{M}$, Retrieval budgets $k_g, k_c$ \\
\textbf{Output:} Generated sequence $Y$
\begin{algorithmic}[1]
\State \textcolor{gray}{// \textbf{Phase 1: Index Construction (Prefill)}}
\State $\mathcal{S} \leftarrow \text{StructureAwareChunking}(X)$
\State $\mathcal{I} \leftarrow \text{HierarchicalIndexing}(\mathcal{S})$ \Comment{Build Coarse $\mathcal{G}$ \& Fine $\mathcal{C}$}
\State $B \leftarrow \emptyset$ \Comment{Initialize token buffer}

\State \textcolor{gray}{// \textbf{Phase 2: Decoding \& Retrieval}}
\For{$t = 1, \dots, T$}
    \State $q_t \leftarrow \mathcal{M}.\text{GetQuery}(x_t)$
    
    \State \textcolor{gray}{// Step 1: Coarse-Level Pruning (Top-$k_g$)}
    \State $\text{Scores}_g \leftarrow \{ q_t^\top \mu_g + \|q_t\| \cdot r_g \mid g \in \mathcal{G} \}$ \Comment{Eqn. 2}
    \State $\mathcal{G}_{top} \leftarrow \text{TopK}(\text{Scores}_g, k_g)$
    
    \State \textcolor{gray}{// Step 2: Fine-Level Pruning (Top-$k_c$)}
    \State $\mathcal{C}_{cand} \leftarrow \bigcup_{g \in \mathcal{G}_{top}} g.\text{children}$
    \State $\text{Scores}_c \leftarrow \{ q_t^\top \mu_c + \|q_t\| \cdot r_c \mid c \in \mathcal{C}_{cand} \}$
    \State $\mathcal{C}_{top} \leftarrow \text{TopK}(\text{Scores}_c, k_c)$
    
    \State \textcolor{gray}{// Step 3: Exact Attention}
    \State $KV_{active} \leftarrow \bigcup_{c \in \mathcal{C}_{top}} c.\text{chunks}.\text{KV}$
    \State $x_{t+1} \leftarrow \text{Attention}(q_t, KV_{active})$
    
    \State \textcolor{gray}{// Step 4: Incremental Update}
    \State $B.\text{append}(\mathcal{M}.\text{GetKV}(x_t))$
    \If{$|B| \ge \text{ChunkSize}$}
        \State $s_{new} \leftarrow \text{Pack}(B); \quad B \leftarrow \emptyset$
        \State $\text{LazyUpdate}(\mathcal{I}, s_{new})$ \Comment{Assign \& expand radius}
    \EndIf
\EndFor
\end{algorithmic}
\end{algorithm}

\subsection{Overall Framework}
\label{subsec:framework}

As illustrated in Figure~\ref{fig:framework}, the \texttt{\model} framework operates in two distinct phases: \textbf{(1) index construction (prefill phase).} When receiving long-context inputs, we do not store a flat KV cache. Instead, we employ a structure-aware algorithm to segment the context, organizing it into a three-level index: coarse units $\to$ fine clusters $\to$ chunks. \textbf{(2) retrieval \& update (decoding phase).} During token generation, the query $q_t$ traverses the index tree top-down. By calculating the upper bound by Eqn.~\ref{eq:upper_bound}, the system rapidly identifies the most relevant KV chunks. Concurrently, newly generated tokens are managed via a lightweight incremental mechanism to maintain index freshness without incurring heavy re-clustering overheads. We provide the details of \texttt{\model} in Algorithm ~\ref{alg:inference}.

\subsection{Prefill Phase}
\label{subsec:index_construction}

In the prefill phase, we process the full prompt context to build the retrieval-ready structure. This involves two key components: structure-aware chunking and hierarchical KV indexing.

\paragraph{Structure-Aware Chunking.} 
Standard fixed size window approaches often disrupt semantic dependencies or syntactic structures, degrading representation quality. To mitigate this, we propose a boundary-aware segmentation algorithm. The algorithm accumulates tokens greedily and, upon reaching a minimum length threshold, looks ahead for high-priority natural delimiters (e.g., paragraph breaks, sentence endings, punctuation). If the maximum threshold is reached without a natural break, a forced split is applied. This process maps the context into a sequence of variable-length, semantically self-contained chunks $S = \{s_1, s_2, ..., s_M\}$.

\paragraph{Hierarchical KV Indexing.} 
To enable logarithmic time retrieval, we organize the chunks into a pyramid structure. The process begins at the chunk level, where we compute the representation $\bar{k}_i$ for each chunk $s_i$ by performing mean pooling on its internal token keys followed by $L_{2}$ normalization. Building upon these representations, we further proceed to the fine cluster level by applying spherical $k$-means clustering~\cite{hornik2012spherical} to partition $\{\bar{k}_i\}$ into $L$ fine-grained clusters. Each cluster $c_j$ is characterized by its centroid $\mu_{c_j}$ and a covering radius $r_{c_j}$, defined as the maximum Euclidean distance from the centroid to any member chunk. Finally, to address scenarios with extremely long contexts where $L$ remains prohibitively large, we construct a coarse unit level. We employ the same logic to aggregate the $L$ cluster centroids into $P$ coarse units ($P \ll L$). This three-tier architecture ($\text{coarse} \to \text{fine} \to \text{chunk}$) drastically reduces the search space for the subsequent decoding phase.

\subsection{Decoding Phase}
\label{subsec:retrieval_update}

During decoding, the system must locate high-relevance caches with sub-linear complexity while accommodating the growing context.

\paragraph{Top-Down Pruning.} 
For a query $q_t$ at each step, we perform a hierarchical search. We first evaluate the upper bound (Eqn.~\ref{eq:upper_bound}) against the top-level coarse units. Specifically, we compute $\text{UB}(q_t, g_p) = q_t^\top \mu_{g_p} + \|q_t\|_2 \cdot r_{g_p}$ and retain only the top-$k_g$ units. Within these selected units, we apply the same logic to rank and select the top-$k_c$ fine clusters. Finally, all KV chunks belonging to the selected fine clusters are loaded as the active set $(K_{\text{active}}, V_{\text{active}})$ for exact attention computation. This \textit{prune-and-refine} strategy ensures that the computational cost remains nearly constant regardless of total context length.

\paragraph{Incremental Update.} 
As new tokens are generated, their corresponding KV cache pairs must be indexed. To avoid the prohibitive cost of global re-clustering, \texttt{\model} adopts a lazy update strategy. During decoding, new tokens are temporarily stored in a buffer. Once enough tokens have accumulated in the buffer to form a full \textit{dynamic chunk}, the chunk is assigned to the nearest existing fine cluster and coarse unit based on centroid proximity. The centroids of the affected nodes are updated via moving averages, while the radii undergo monotonic expansion to encompass the new chunk. This design allows the model to handle infinite streaming generation while maintaining the integrity of the retrieval index.

\section{Experiments}
    \label{sec:experiments}
\subsection{Experimental Settings}

\paragraph{Benchmarks and Models.} 
To evaluate the performance over different scenarios, we conduct experiments across models of varying architectures and sizes. For long-context understanding, we benchmark the Llama-3.1-8B-Instruct~\cite{LlamaTeam2024llama3} model on the \textsc{LongBench V2}~\cite{bai2025longbench} dataset. It contains complex real world long-context questions across diverse tasks such as long in-context learning, long structured data analysis, and code repository understanding. With context lengths ranging from 8K to 2M, \textsc{LongBench V2} emphasizes deep reasoning and noise resistance, distinguishing it from LongBench~\cite{bai2024longbench} which focuses on simple retrieval. This allows for rigorous testing of semantic understanding over extremely long sequences. For complex reasoning scenarios, we employ the Deepseek-R1-Distill-Llama-8B and Deepseek-R1-Distill-Qwen-14B~\cite{deepseekai2025deepseekr1} models, evaluating them on the \textsc{MATH500}~\cite{Dan2021math500} dataset. It consists of challenging mathematics competition problems and is widely used to assess long chain-of-thought reasoning performance. Due to space limitations, results of \textsc{RULER} benchmark~\cite{hsieh2024ruler} on Appendix~\ref{app:ruler}.

\paragraph{Baseline.} 
We compare \texttt{\model} against full attention and several state-of-the-art KV management and sparse attention methods including Quest~\cite{quest}, ClusterKV~\cite{clusterkv}, ArkVale~\cite{renze2024arkvale}, RaaS~\cite{hu2025raas}, ShadowKV~\cite{sun2024shadowkv}, and RazorAttention~\cite{tang2024razorattention}. Since ClusterKV is equivalent to full attention when context length does not exceed the budget, and the context lengths in MATH500 are relatively short, we exclude it from complex reasoning experiments.
The implementation details are in Appendix~\ref{appendix:implement}.

\subsection{Performance Evaluation}
\begin{table*}[t!]
\centering
\small
\renewcommand{\arraystretch}{1.2} 
\setlength{\tabcolsep}{18pt} 

\begin{tabular}{lcccc}
\toprule
\textbf{Method} & \textbf{Overall} & \textbf{Short} & \textbf{Medium} & \textbf{Long} \\
\midrule
Full Attention  & 30.02 & 35.56 & 27.91 & 25.00  \\
RazorAttention~\cite{tang2024razorattention}  & 27.44 & 33.89 & 25.12 & 21.30  \\
RaaS~\cite{hu2025raas}            & 28.23 & 33.89 & 26.51 & 22.02  \\
ArkVale~\cite{renze2024arkvale}         & 28.63 & 33.89 & 26.98 & 23.15  \\
ShadowKV~\cite{sun2024shadowkv}        & 25.45 & 32.78 & 22.79 & 18.52  \\
Quest~\cite{quest}           & 20.68 & 22.78 & 19.07 & 20.37  \\
ClusterKV~\cite{clusterkv}       & 26.64 & 30.56 & 24.65 & \textbf{24.07}  \\
\midrule
\textbf{\model} & \textbf{\textcolor{blue}{30.82}} & \textbf{\textcolor{blue}{37.22}} & \textbf{\textcolor{blue}{28.84}} & \textbf{24.07} \\
\bottomrule
\end{tabular}
\caption{Performance comparison on LongBench V2 using Llama-3.1-8B-Instruct. \textbf{Bold} denotes the best sparse method; \textcolor{blue}{Blue} highlights cases where \model outperforms Full Attention.}
\vspace{-4mm}
\label{tab:longbench-v2}
\end{table*}

\paragraph{Performance on Long-Context Understanding.}
As shown in Table~\ref{tab:longbench-v2}, the results indicate that on the LongBench V2 benchmark, \texttt{\model} achieves an overall accuracy of 30.8\% with a 1024 token budget. The margin compared with other methods validates that preserving semantic integrity via structure-aware chunking is superior to rigid paging or token-level clustering, which often fragment context. Furthermore, the slight improvement over full attention suggests that our hierarchical pruning effectively acts as a noise filter, enabling the model to focus on critical information within noisy long contexts.


\paragraph{Performance on Complex Reasoning.}
We evaluate the reasoning capability of \texttt{\model} on the MATH500, results shown in Table~\ref{tab:math500}. Complex mathematical reasoning typically involves long chains of thought, where recalling early premises is crucial. \texttt{\model} keeps performance loss within 2\% on Deepseek-R1-Distill-Llama-8B, and surpasses the full attention  on Deepseek-R1-Distill-Qwen-14B. This demonstrates that our method effectively adapts to the dynamic KV cache distribution; even after generating thousands of tokens, it can rapidly retrieve key information from the problem statement and reasoning process, effectively overcoming the bottlenecks of traditional sparse attention in complex reasoning.

\begin{table}[t!]
\centering
\renewcommand{\arraystretch}{1.4} 
\small
\resizebox{\linewidth}{!}{
\begin{tabular}{lcc}
\toprule
\multirow{2}{*}{\textbf{Method}} & \textbf{DeepSeek-R1-Distill} & \textbf{DeepSeek-R1-Distill} \\
 & \textbf{Llama-8B} & \textbf{Qwen-14B} \\
\midrule
Full Attention  & 78.40 & 74.00 \\
RazorAttention  & 72.00 & 70.00 \\
RaaS            & 74.00 & 68.00 \\
ArkVale         & 72.00 & 72.00 \\
ShadowKV        & 76.00 & 76.00 \\
Quest           & 72.00 & 76.00 \\
\bottomrule
\textbf{\model} & \textbf{77.00} & \textbf{74.80} \\
\bottomrule
\end{tabular}
}
\caption{Accuracy on MATH500. \texttt{\model} demonstrates consistent improvements over sparse baselines across different model architectures on DeepSeek-R1-Distill-Llama-8B.}
\vspace{-2mm}
\label{tab:math500}
\end{table}

\subsection{Efficiency Evaluation}

\begin{figure}[t]
    \centering
    \includegraphics[width=\linewidth]{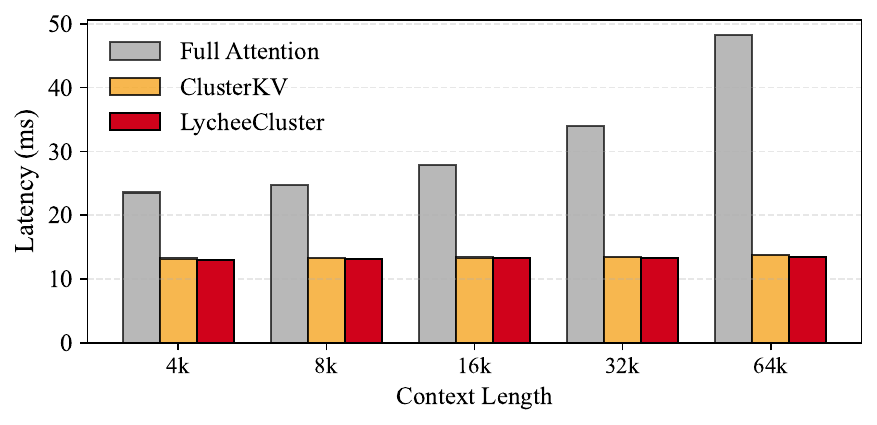} 
    \caption{End-to-end decoding latency (TPOT) comparison on a single H20 GPU across varying context lengths. While full attention exhibits linear latency growth, our method maintains consistently low latency.}
    \vspace{-4mm}
    \label{fig:topt}
\end{figure}

\paragraph{End-to-End Decoding Speedup.}
We use Time Per Output Token (TPOT) as the primary metric to evaluate the decoding latency of \texttt{\model}, comparing it with ClusterKV and full attention based on FlashAttention-2~\cite{flashattention2}) under varying context lengths and token budgets.
As illustrated in Figure~\ref{fig:topt}, the decoding latency of full attention grows linearly with context length. In comparison, \texttt{\model} consistently maintains low latency. At a 32K context length, we achieve a 2.6$\times$ speedup relative to full attention; when the context extends to 64K, this speedup reaches 3.6$\times$. This result confirms the high efficiency of \texttt{\model} when processing long contexts.

\paragraph{Kernel-Level Analysis.}
To gain deeper insights into the sources of efficiency improvements and justify the overhead of clustering and retrieval, we decompose the runtime into four components: (1) index construction (prefill only), (2) hierarchical retrieval, (3) index update and (4) sparse attention.

\begin{figure*}[t]
    \centering
    \includegraphics[width=\linewidth]{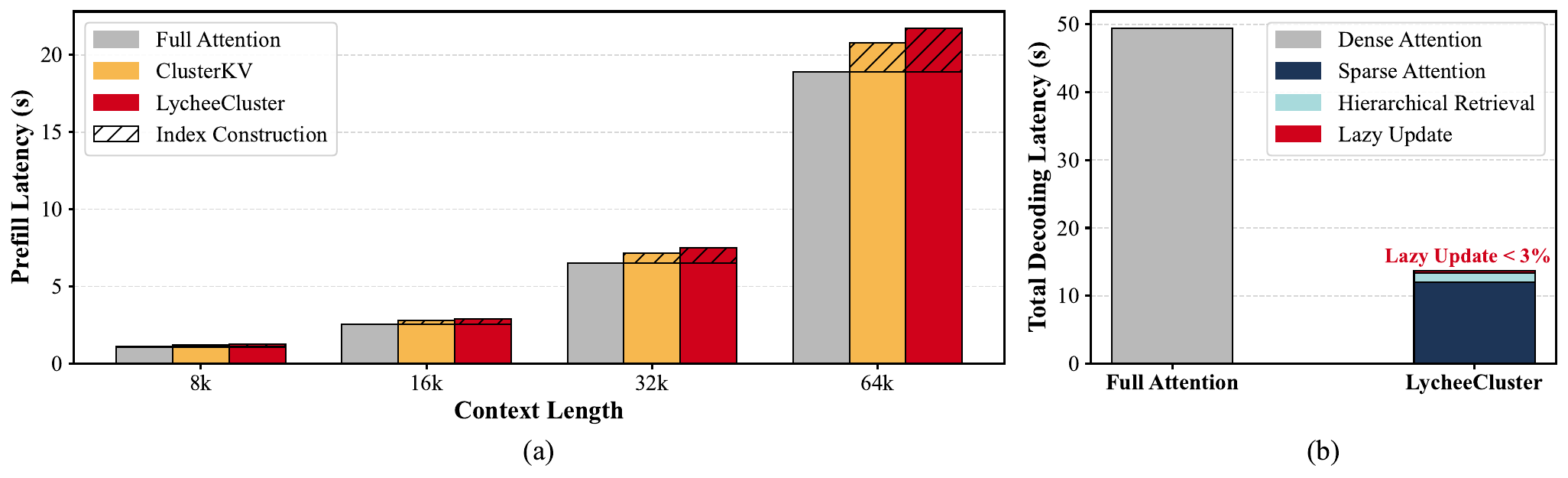} 
    \vspace{-4mm}
    \caption{Kernel-level latency breakdown. \textbf{(a) Prefill Phase:} Latency comparison across varying context lengths. The colored top sections represent the index construction overhead. While \texttt{\model} incurs a slightly higher construction cost (10–15\%) than ClusterKV, it remains a minor fraction of the total prefill time. \textbf{(b) Decoding Phase:} Breakdown of total latency for generating 1,024 tokens at 72k context. The combined overhead of retrieval and lazy updates in \texttt{\model} is minimal compared to the massive computation reduction.}
    \vspace{-4mm}
    \label{fig:knl}
\end{figure*}

Figure~\ref{fig:knl}(a) shows the time breakdown during the prefill phase. Although constructing the hierarchical index introduces additional computation, measurements indicate that this process accounts for only 10\% - 15\% of the total prefill time. Considering that prefill is a one-time operation and subsequent decoding typically involves generating thousands of tokens, its amortized cost is almost negligible.
In contrast, Figure~\ref{fig:knl}(b) breaks down the latency of a single decoding step at a 72K context length, with the total time divided into  hierarchical retrieval, index update and sparse attention. Because of the logarithmic complexity of hierarchical retrieval, the retrieval operation occupies only a minimal fraction of the single-step decoding time; this overhead is far outweighed by the benefits of reduced attention computation. Furthermore, the cost of index maintenance is negligible. Our lazy update strategy consumes less than 1\% of the decoding time. These results indicate that \texttt{\model} achieves high-precision retrieval without disrupting generation fluency.

\subsection{Ablation Study}

\begin{figure}[t]
    \centering
    \includegraphics[width=\linewidth]{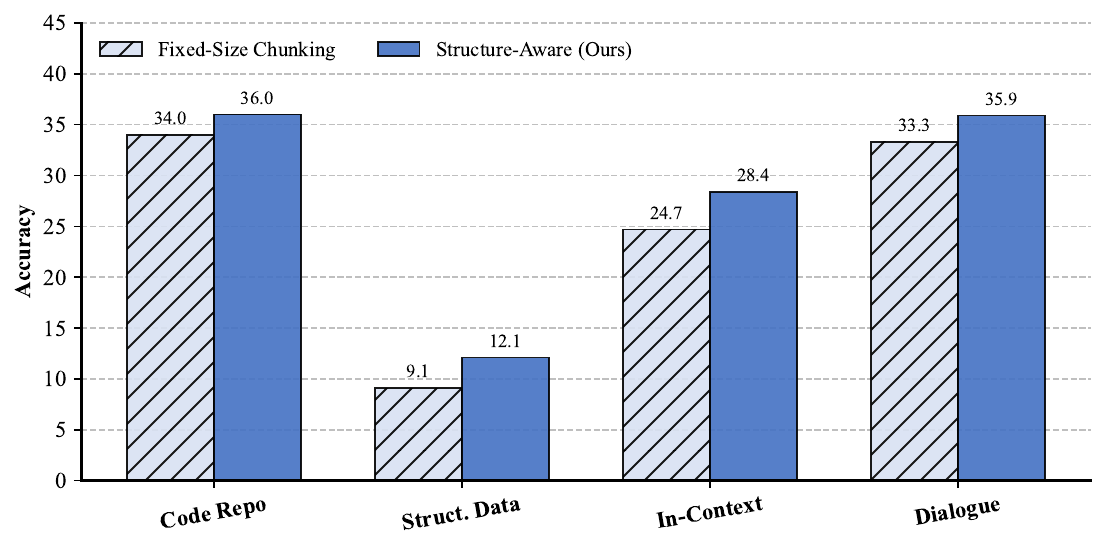}
    \caption{Ablation study on the impact of segmentation strategies. We compare the accuracy of \texttt{\model} using our structure-aware chunking against a fixed-size chunking baseline across four task categories. The results demonstrate that preserving semantic boundaries consistently improves performance.}
    \vspace{-4mm}
    \label{fig:ablation_chunking}
\end{figure}

\paragraph{Impact of Structure-Aware Chunking.}
To assess the impact of our structure-aware chunking design, we replace the structure-aware chunking with fixed-size chunking (chunk size = 16) while keeping the rest of the method unchanged. As illustrated in Figure~\ref{fig:ablation_chunking}, this results in a 3.03\% performance drop for \texttt{\model} on the Long Structured Data Understanding task of LongBench V2, along with a degradation in Code Repository Understanding. This confirms that for semantic retrieval to function effectively, the fundamental unit of retrieval (chunk) must be semantically complete.

\begin{table}[t]
    \centering
    \small
    
    \resizebox{\linewidth}{!}{
        \begin{tabular}{lccccc}
            \toprule
            \textbf{Strategy} & \textbf{Overall} & \textbf{Short} & \textbf{Medium} & \textbf{Long} & \textbf{Recall Rate} \\
            \midrule
            Max & 28.83 & 35.56 & 26.98 & 21.3 & 33.60\% \\
            \textbf{Mean} & \textbf{30.82} & \textbf{37.22} & \textbf{28.84} & \textbf{24.07} & \textbf{40.37\%} \\
            \bottomrule
        \end{tabular}
    }
    \caption{Ablation study on representative key pooling strategies. We compare mean pooling (Ours) against max pooling on the LongBench V2 (Llama-3.1-8B-Instruct).~\textbf{Recall Rate} denotes the retrieval recall rate. It is defined as the percentage of the top-$k$ tokens with the highest ground-truth attention scores (computed by full attention) that are successfully retrieved by the strategy within the same token budget.}
    \vspace{-4mm}
    \label{tab:pooling_ablation}
    
\end{table}

\paragraph{Pooling Strategy for Chunk Representation.}
We investigate the impact of the aggregation function used to generate the representative key $\bar{k}_j$ for each chunk. We compare our mean pooling strategy against max pooling. The results are illustrated in Table~\ref{tab:pooling_ablation}, mean pooling consistently outperforms max pooling across all subsets. We attribute this to the geometric nature of the self-attention mechanism. Since attention scores are derived from dot products (measuring alignment between Query and Key vectors), preserving the directionality of the feature space is crucial. Max pooling constructs a synthetic vector composed of extreme values from each dimension, which often distorts the original semantic direction and is sensitive to outliers. In contrast, mean pooling followed by $L_2$  normalization effectively computes the geometric centroid of the token embeddings of the chunk. This provides a faithful representation of the  average semantic position of the chunk, mathematically aligning with our spherical $k$-means objective to maximize the retrieval recall of relevant semantic information.

\begin{figure}[t]
    \centering
    \includegraphics[width=\linewidth]{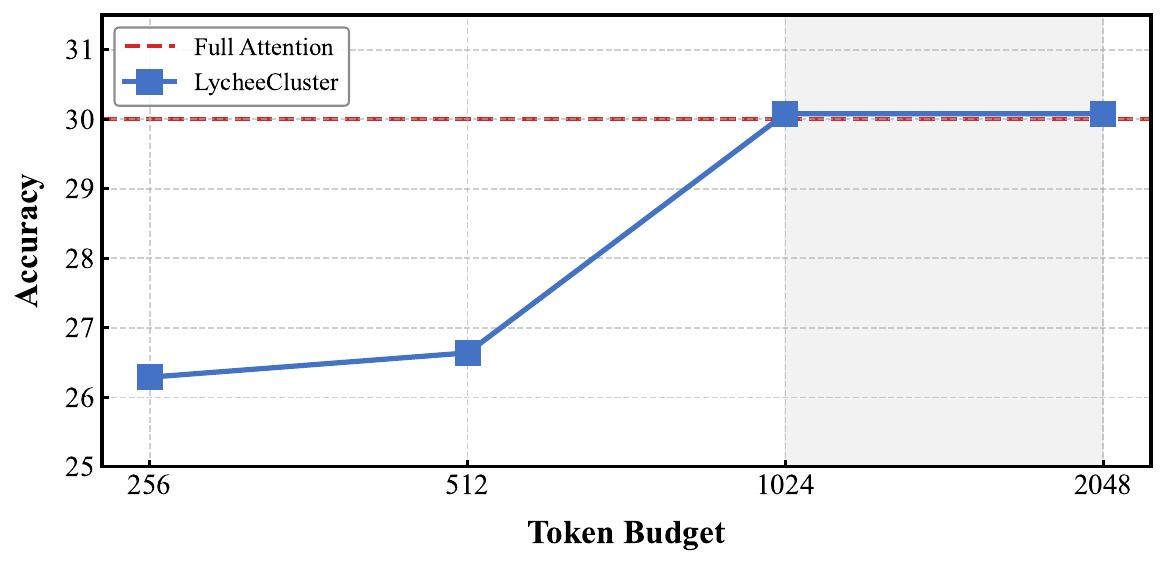}
    \caption{Impact of token budget on LongBench V2 performance. The accuracy improves as the budget increases from 256 to 1024.}
    \vspace{-4mm}
    \label{fig:ablation_token}
\end{figure}

\paragraph{Impact of Token Budget.}
We analyze the impact of the token budget on performance. As shown in Figure~\ref{fig:ablation_token}, overall accuracy in LongBench V2 improves significantly as the budget increases from 256 to 1024, but then saturates, showing limited gains with larger budgets. This demonstrates that \texttt{\model} can effectively utilize a small budget to precisely capture the critical tokens within extremely long contexts that are essential for generation, thereby maintaining model performance.

\section{Conclusion}

We proposed \texttt{\model} that harmonizes memory efficiency with semantic integrity. By employing structure-aware chunking and  hierarchical index, we achieve bounded, sub-linear retrieval without disrupting local context. Our approach outperforms state-of-the-art methods, delivering comparable accuracy to full attention with lower latency. It offers a scalable solution for long-context LLMs inference in resource-constrained scenarios.

\section*{Limitations}


First, one limitation of our method is that we focus on speeding up generation, and the proposed method does not accelerate attention during the prefill phase. However, prefill overhead is a one-time cost. In long-context scenarios (e.g., RAG, agentic workflows), the decoding phase (generation) typically dominates the total latency. The 3.6x speedup in decoding quickly offsets the prefill cost. One promising direction for future research would be to extend our method to support accelerating prefill. Second, integrating our proposed method into established LLM serving frameworks (e.g., vLLM~\cite{kwon2023vllm}, SGLang~\cite{zheng2024sglang}) necessitates additional engineering effort.  Furthermore, more adaptive strategies can further enhance the generalization ability of this method. As part of ongoing and future work, we are actively investigating the relationship between input features (e.g., sentence length distribution and semantic diversity) and the optimal semantic-aware chunking factor. Preliminary observations suggest that dynamically adjusting this factor based on these input statistics can further improve efficiency and accuracy. Finally, while our technique incurs a memory overhead for storing centroids although this cost is negligible when compared to the total size of the KV cache.

\section*{Ethical Considerations}
Our work primarily contributes to Green AI by reducing the inference energy consumption of LLMs through sub-linear retrieval complexity. This efficiency enables the deployment of long-context models on resource-constrained hardware, democratizing access to advanced AI. However, as an approximate attention mechanism, there is a theoretical risk of information loss in safety-critical domains (e.g., medical or legal analysis), necessitating careful verification in high-stakes applications. Finally, while efficient inference lowers barriers for beneficial use, it could also facilitate large-scale misuse by malicious actors; we advocate for responsible deployment to mitigate these risks.
\bibliography{custom}

\begin{thebibliography}{36}
\providecommand{\natexlab}[1]{#1}

\bibitem[{Anthropic(2025)}]{claude-3.7}
Anthropic. 2025.
\newblock Claude 3.7 sonnet system card.

\bibitem[{Bai et~al.(2024)Bai, Lv, Zhang, Lyu, Tang, Huang, Du, Liu, Zeng, Hou, Dong, Tang, and Li}]{bai2024longbench}
Yushi Bai, Xin Lv, Jiajie Zhang, Hongchang Lyu, Jiankai Tang, Zhidian Huang, Zhengxiao Du, Xiao Liu, Aohan Zeng, Lei Hou, Yuxiao Dong, Jie Tang, and Juanzi Li. 2024.
\newblock \href {https://doi.org/10.18653/V1/2024.ACL-LONG.172} {Longbench: {A} bilingual, multitask benchmark for long context understanding}.
\newblock In \emph{Proceedings of the 62nd Annual Meeting of the Association for Computational Linguistics (Volume 1: Long Papers), {ACL} 2024, Bangkok, Thailand, August 11-16, 2024}, pages 3119--3137. Association for Computational Linguistics.

\bibitem[{Bai et~al.(2025)Bai, Tu, Zhang, Peng, Wang, Lv, Cao, Xu, Hou, Dong et~al.}]{bai2025longbench}
Yushi Bai, Shangqing Tu, Jiajie Zhang, Hao Peng, Xiaozhi Wang, Xin Lv, Shulin Cao, Jiazheng Xu, Lei Hou, Yuxiao Dong, and 1 others. 2025.
\newblock Longbench v2: Towards deeper understanding and reasoning on realistic long-context multitasks.
\newblock In \emph{Proceedings of the 63rd Annual Meeting of the Association for Computational Linguistics (Volume 1: Long Papers)}, pages 3639--3664.

\bibitem[{Chen et~al.(2025)Chen, Liu, Xu, Gao, and Wang}]{chen2025sablock}
Jinhan Chen, Jianchun Liu, Hongli Xu, Xianjun Gao, and Shilong Wang. 2025.
\newblock Sablock: Semantic-aware kv cache eviction with adaptive compression block size.
\newblock \emph{arXiv preprint arXiv:2510.22556}.

\bibitem[{Chen et~al.(2024)Chen, Wang, Cao, Wu, Zheng, Li, Wei, Yan, Li, and Liang}]{renze2024arkvale}
Renze Chen, Zhuofeng Wang, Beiquan Cao, Tong Wu, Size Zheng, Xiuhong Li, Xuechao Wei, Shengen Yan, Meng Li, and Yun Liang. 2024.
\newblock \href {http://papers.nips.cc/paper\_files/paper/2024/hash/cd4b49379efac6e84186a3ffce108c37-Abstract-Conference.html} {Arkvale: Efficient generative {LLM} inference with recallable key-value eviction}.
\newblock In \emph{Advances in Neural Information Processing Systems 38: Annual Conference on Neural Information Processing Systems 2024, NeurIPS 2024, Vancouver, BC, Canada, December 10 - 15, 2024}.

\bibitem[{Comanici et~al.(2025)Comanici, Bieber, Schaekermann, Pasupat, Sachdeva, Dhillon, Blistein, Ram, Zhang, Rosen et~al.}]{comanici2025gemini}
Gheorghe Comanici, Eric Bieber, Mike Schaekermann, Ice Pasupat, Noveen Sachdeva, Inderjit Dhillon, Marcel Blistein, Ori Ram, Dan Zhang, Evan Rosen, and 1 others. 2025.
\newblock Gemini 2.5: Pushing the frontier with advanced reasoning, multimodality, long context, and next generation agentic capabilities.
\newblock \emph{arXiv preprint arXiv:2507.06261}.

\bibitem[{Dao(2024)}]{flashattention2}
Tri Dao. 2024.
\newblock \href {https://openreview.net/forum?id=mZn2Xyh9Ec} {Flashattention-2: Faster attention with better parallelism and work partitioning}.
\newblock In \emph{The Twelfth International Conference on Learning Representations, {ICLR} 2024, Vienna, Austria, May 7-11, 2024}.

\bibitem[{DeepSeek{-}AI(2025)}]{deepseekai2025deepseekr1}
DeepSeek{-}AI. 2025.
\newblock \href {https://doi.org/10.48550/ARXIV.2501.12948} {Deepseek-r1: Incentivizing reasoning capability in llms via reinforcement learning}.
\newblock \emph{CoRR}, abs/2501.12948.

\bibitem[{Fountas et~al.(2025)Fountas, Benfeghoul, Oomerjee, Christopoulou, Lampouras, Ammar, and Wang}]{fountas2025human}
Zafeirios Fountas, Martin Benfeghoul, Adnan Oomerjee, Fenia Christopoulou, Gerasimos Lampouras, Haitham~Bou Ammar, and Jun Wang. 2025.
\newblock Human-inspired episodic memory for infinite context llms.
\newblock In \emph{The Thirteenth International Conference on Learning Representations}.

\bibitem[{Gu et~al.(2025)Gu, Ye, Chen, Zhou, Feng, and Xiao}]{gu2025structext}
Zhouhong Gu, Haoning Ye, Xingzhou Chen, Zeyang Zhou, Hongwei Feng, and Yanghua Xiao. 2025.
\newblock Structext-eval: Evaluating large language model’s reasoning ability in structure-rich text.
\newblock In \emph{Proceedings of the 63rd Annual Meeting of the Association for Computational Linguistics (Volume 1: Long Papers)}, pages 223--244.

\bibitem[{Hendrycks et~al.(2021{\natexlab{a}})Hendrycks, Burns, Kadavath, Arora, Basart, Tang, Song, and Steinhardt}]{Dan2021math500}
Dan Hendrycks, Collin Burns, Saurav Kadavath, Akul Arora, Steven Basart, Eric Tang, Dawn Song, and Jacob Steinhardt. 2021{\natexlab{a}}.
\newblock \href {https://datasets-benchmarks-proceedings.neurips.cc/paper/2021/hash/be83ab3ecd0db773eb2dc1b0a17836a1-Abstract-round2.html} {Measuring mathematical problem solving with the {MATH} dataset}.
\newblock In \emph{Proceedings of the Neural Information Processing Systems Track on Datasets and Benchmarks 1, NeurIPS Datasets and Benchmarks 2021, December 2021, virtual}.

\bibitem[{Hendrycks et~al.(2021{\natexlab{b}})Hendrycks, Burns, Kadavath, Arora, Basart, Tang, Song, and Steinhardt}]{hendrycks2021measuring}
Dan Hendrycks, Collin Burns, Saurav Kadavath, Akul Arora, Steven Basart, Eric Tang, Dawn Song, and Jacob Steinhardt. 2021{\natexlab{b}}.
\newblock Measuring mathematical problem solving with the math dataset.
\newblock \emph{NeurIPS}.

\bibitem[{Hornik et~al.(2012)Hornik, Feinerer, Kober, and Buchta}]{hornik2012spherical}
Kurt Hornik, Ingo Feinerer, Martin Kober, and Christian Buchta. 2012.
\newblock Spherical k-means clustering.
\newblock \emph{Journal of statistical software}, 50:1--22.

\bibitem[{Hsieh et~al.(2024)Hsieh, Sun, Kriman, Acharya, Rekesh, Jia, and Ginsburg}]{hsieh2024ruler}
Cheng-Ping Hsieh, Simeng Sun, Samuel Kriman, Shantanu Acharya, Dima Rekesh, Fei Jia, and Boris Ginsburg. 2024.
\newblock \href {https://openreview.net/forum?id=kIoBbc76Sy} {{RULER}: What{\textquoteright}s the real context size of your long-context language models?}
\newblock In \emph{First Conference on Language Modeling}.

\bibitem[{Hu et~al.(2025)Hu, Huang, Wang, Li, Hu, Liu, Chen, Xie, and Shan}]{hu2025raas}
Junhao Hu, Wenrui Huang, Weidong Wang, Zhenwen Li, Tiancheng Hu, Zhixia Liu, Xusheng Chen, Tao Xie, and Yizhou Shan. 2025.
\newblock \href {https://aclanthology.org/2025.findings-acl.131/} {Raas: Reasoning-aware attention sparsity for efficient {LLM} reasoning}.
\newblock In \emph{Findings of the Association for Computational Linguistics, {ACL} 2025, Vienna, Austria, July 27 - August 1, 2025}, pages 2577--2590.

\bibitem[{Kim et~al.(2025)Kim, Kim, Kwon, Lee, Yun, and Song}]{kvzip}
Jang{-}Hyun Kim, Jinuk Kim, Sangwoo Kwon, Jae~W. Lee, Sangdoo Yun, and Hyun~Oh Song. 2025.
\newblock \href {https://doi.org/10.48550/ARXIV.2505.23416} {Kvzip: Query-agnostic {KV} cache compression with context reconstruction}.
\newblock \emph{CoRR}, abs/2505.23416.

\bibitem[{Kwon et~al.(2023)Kwon, Li, Zhuang, Sheng, Zheng, Yu, Gonzalez, Stoica, and Wu}]{kwon2023vllm}
Woosuk Kwon, Zhuohan Li, Siyuan Zhuang, Ying Sheng, Lianmin Zheng, Cody~Hao Yu, Joseph~E. Gonzalez, Ion Stoica, and Zhanghao Wu. 2023.
\newblock Efficient memory management for large language model serving with pagedattention.
\newblock In \emph{Proceedings of the 29th Symposium on Operating Systems Principles}, pages 611--626.

\bibitem[{Li et~al.(2024{\natexlab{a}})Li, Li, Tian, Tang, Xu, Chen, Hu, Dong, Li, and Chen}]{li2024survey}
Haoyang Li, Yiming Li, Anxin Tian, Tianhao Tang, Zhanchao Xu, Xuejia Chen, Nicole Hu, Wei Dong, Qing Li, and Lei Chen. 2024{\natexlab{a}}.
\newblock A survey on large language model acceleration based on kv cache management.
\newblock \emph{arXiv preprint arXiv:2412.19442}.

\bibitem[{Li et~al.(2025)Li, Jiang, Wu, Luo, Ahn, Zhang, Abdi, Li, Gao, Yang et~al.}]{liscbench}
Yucheng Li, Huiqiang Jiang, Qianhui Wu, Xufang Luo, Surin Ahn, Chengruidong Zhang, Amir~H Abdi, Dongsheng Li, Jianfeng Gao, Yuqing Yang, and 1 others. 2025.
\newblock Scbench: A kv cache-centric analysis of long-context methods.
\newblock In \emph{The Thirteenth International Conference on Learning Representations}.

\bibitem[{Li et~al.(2024{\natexlab{b}})Li, Huang, Yang, Venkitesh, Locatelli, Ye, Cai, Lewis, and Chen}]{snapkv}
Yuhong Li, Yingbing Huang, Bowen Yang, Bharat Venkitesh, Acyr Locatelli, Hanchen Ye, Tianle Cai, Patrick Lewis, and Deming Chen. 2024{\natexlab{b}}.
\newblock \href {http://papers.nips.cc/paper\_files/paper/2024/hash/28ab418242603e0f7323e54185d19bde-Abstract-Conference.html} {Snapkv: {LLM} knows what you are looking for before generation}.
\newblock In \emph{Advances in Neural Information Processing Systems 38: Annual Conference on Neural Information Processing Systems 2024, NeurIPS 2024, Vancouver, BC, Canada, December 10 - 15, 2024}.

\bibitem[{Lightman et~al.(2023)Lightman, Kosaraju, Burda, Edwards, Baker, Lee, Leike, Schulman, Sutskever, and Cobbe}]{lightman2023let}
Hunter Lightman, Vineet Kosaraju, Yuri Burda, Harrison Edwards, Bowen Baker, Teddy Lee, Jan Leike, John Schulman, Ilya Sutskever, and Karl Cobbe. 2023.
\newblock Let's verify step by step.
\newblock In \emph{The Twelfth International Conference on Learning Representations}.

\bibitem[{Liu et~al.(2024)Liu, Chen, Lu, Jiang, Han, Zhang, Chen, Zhang, Ding, Zhang et~al.}]{liu2024retrievalattention}
Di~Liu, Meng Chen, Baotong Lu, Huiqiang Jiang, Zhenhua Han, Qianxi Zhang, Qi~Chen, Chengruidong Zhang, Bailu Ding, Kai Zhang, and 1 others. 2024.
\newblock Retrievalattention: Accelerating long-context llm inference via vector retrieval.
\newblock \emph{arXiv preprint arXiv:2409.10516}.

\bibitem[{Liu et~al.(2025{\natexlab{a}})Liu, Li, Zhao, Zhang, and Guo}]{clusterkv}
Guangda Liu, Chengwei Li, Jieru Zhao, Chenqi Zhang, and Minyi Guo. 2025{\natexlab{a}}.
\newblock \href {https://doi.org/10.1109/DAC63849.2025.11132479} {Clusterkv: Manipulating {LLM} {KV} cache in semantic space for recallable compression}.
\newblock In \emph{62nd {ACM/IEEE} Design Automation Conference, {DAC} 2025, San Francisco, CA, USA, June 22-25, 2025}, pages 1--7. {IEEE}.

\bibitem[{Liu et~al.(2025{\natexlab{b}})Liu, Zhu, Bai, He, Liao, Que, Wang, Zhang, Zhang, Zhang et~al.}]{liu2025comprehensive}
Jiaheng Liu, Dawei Zhu, Zhiqi Bai, Yancheng He, Huanxuan Liao, Haoran Que, Zekun Wang, Chenchen Zhang, Ge~Zhang, Jiebin Zhang, and 1 others. 2025{\natexlab{b}}.
\newblock A comprehensive survey on long context language modeling.
\newblock \emph{arXiv preprint arXiv:2503.17407}.

\bibitem[{Lu et~al.(2025)Lu, Jiang, Liu, Du, Jiang, Hong, Liu, He, Yuan, Wang, Huang, Yuan, Xu, Xu, Lai, Chen, Zheng, Yan, Su, Wu, Zhang, Yang, Zhou, Zhang, and Qiu}]{lu2025moba}
Enzhe Lu, Zhejun Jiang, Jingyuan Liu, Yulun Du, Tao Jiang, Chao Hong, Shaowei Liu, Weiran He, Enming Yuan, Yuzhi Wang, Zhiqi Huang, Huan Yuan, Suting Xu, Xinran Xu, Guokun Lai, Yanru Chen, Huabin Zheng, Junjie Yan, Jianlin Su, and 6 others. 2025.
\newblock \href {https://openreview.net/forum?id=RlqYCpTu1P} {Mo{BA}: Mixture of block attention for long-context {LLM}s}.
\newblock In \emph{The Thirty-ninth Annual Conference on Neural Information Processing Systems}.

\bibitem[{Sun et~al.(2025{\natexlab{a}})Sun, Chang, Bao, Zheng, Zheng, Liu, Dong, Chi, and Chen}]{sun2024shadowkv}
Hanshi Sun, Li-Wen Chang, Wenlei Bao, Size Zheng, Ningxin Zheng, Xin Liu, Harry Dong, Yuejie Chi, and Beidi Chen. 2025{\natexlab{a}}.
\newblock \href {https://openreview.net/forum?id=oa7MYAO6h6} {Shadow{KV}: {KV} cache in shadows for high-throughput long-context {LLM} inference}.
\newblock In \emph{Forty-second International Conference on Machine Learning}.

\bibitem[{Sun et~al.(2025{\natexlab{b}})Sun, Ye, Dong, Xia, Chen, Gao, Cao, Wang, and Wei}]{sun2025rectified}
Yutao Sun, Tianzhu Ye, Li~Dong, Yuqing Xia, Jian Chen, Yizhao Gao, Shijie Cao, Jianyong Wang, and Furu Wei. 2025{\natexlab{b}}.
\newblock Rectified sparse attention.
\newblock \emph{arXiv preprint arXiv:2506.04108}.

\bibitem[{Tang et~al.(2025)Tang, Lin, Lin, Han, Ke, Hong, Yao, and Wang}]{tang2024razorattention}
Hanlin Tang, Yang Lin, Jing Lin, Qingsen Han, Danning Ke, Shikuan Hong, Yiwu Yao, and Gongyi Wang. 2025.
\newblock Razorattention: Efficient kv cache compression through retrieval heads.
\newblock In \emph{The Thirteenth International Conference on Learning Representations}.

\bibitem[{Tang et~al.(2024)Tang, Zhao, Zhu, Xiao, Kasikci, and Han}]{quest}
Jiaming Tang, Yilong Zhao, Kan Zhu, Guangxuan Xiao, Baris Kasikci, and Song Han. 2024.
\newblock \href {https://openreview.net/forum?id=KzACYw0MTV} {{QUEST:} query-aware sparsity for efficient long-context {LLM} inference}.
\newblock In \emph{Forty-first International Conference on Machine Learning, {ICML} 2024, Vienna, Austria, July 21-27, 2024}. OpenReview.net.

\bibitem[{Team(2024)}]{LlamaTeam2024llama3}
Llama Team. 2024.
\newblock \href {https://doi.org/10.48550/ARXIV.2407.21783} {The llama 3 herd of models}.
\newblock \emph{CoRR}, abs/2407.21783.

\bibitem[{Xiao et~al.(2024)Xiao, Tian, Chen, Han, and Lewis}]{streamingllm}
Guangxuan Xiao, Yuandong Tian, Beidi Chen, Song Han, and Mike Lewis. 2024.
\newblock \href {https://openreview.net/forum?id=NG7sS51zVF} {Efficient streaming language models with attention sinks}.
\newblock In \emph{The Twelfth International Conference on Learning Representations, {ICLR} 2024, Vienna, Austria, May 7-11, 2024}. OpenReview.net.

\bibitem[{Yang et~al.(2025{\natexlab{a}})Yang, Li, Yang, Zhang, Hui, Zheng, Yu, Gao, Huang, Lv et~al.}]{qwen3_technical_report}
An~Yang, Anfeng Li, Baosong Yang, Beichen Zhang, Binyuan Hui, Bo~Zheng, Bowen Yu, Chang Gao, Chengen Huang, Chenxu Lv, and 1 others. 2025{\natexlab{a}}.
\newblock Qwen3 technical report.
\newblock \emph{arXiv preprint arXiv:2505.09388}.

\bibitem[{Yang et~al.(2025{\natexlab{b}})Yang, Zhang, Chen, Li, and Jia}]{tidaldecode}
Lijie Yang, Zhihao Zhang, Zhuofu Chen, Zikun Li, and Zhihao Jia. 2025{\natexlab{b}}.
\newblock \href {https://openreview.net/forum?id=EkfLaCJ7bk} {Tidaldecode: Fast and accurate {LLM} decoding with position persistent sparse attention}.
\newblock In \emph{The Thirteenth International Conference on Learning Representations, {ICLR} 2025, Singapore, April 24-28, 2025}. OpenReview.net.

\bibitem[{Zhang et~al.(2023)Zhang, Sheng, Zhou, Chen, Zheng, Cai, Song, Tian, R{\'{e}}, Barrett, Wang, and Chen}]{h2o}
Zhenyu Zhang, Ying Sheng, Tianyi Zhou, Tianlong Chen, Lianmin Zheng, Ruisi Cai, Zhao Song, Yuandong Tian, Christopher R{\'{e}}, Clark~W. Barrett, Zhangyang Wang, and Beidi Chen. 2023.
\newblock \href {http://papers.nips.cc/paper\_files/paper/2023/hash/6ceefa7b15572587b78ecfcebb2827f8-Abstract-Conference.html} {{H2O:} heavy-hitter oracle for efficient generative inference of large language models}.
\newblock In \emph{Advances in Neural Information Processing Systems 36: Annual Conference on Neural Information Processing Systems 2023, NeurIPS 2023, New Orleans, LA, USA, December 10 - 16, 2023}.

\bibitem[{Zheng et~al.(2024)Zheng, Yin, Xie, Sun, Huang, Yu, Cao, Kozyrakis, Stoica, Gonzalez et~al.}]{zheng2024sglang}
Lianmin Zheng, Liangsheng Yin, Zhiqiang Xie, Chuyue~Livia Sun, Jeff Huang, Cody~Hao Yu, Shiyi Cao, Christos Kozyrakis, Ion Stoica, Joseph~E Gonzalez, and 1 others. 2024.
\newblock Sglang: Efficient execution of structured language model programs.
\newblock \emph{Advances in neural information processing systems}, 37:62557--62583.

\bibitem[{Zhu et~al.(2025)Zhu, Falahati, Yang, and Amiri}]{zhu2025sentencekv}
Yuxuan Zhu, Ali Falahati, David~H. Yang, and Mohammad~Mohammadi Amiri. 2025.
\newblock \href {https://openreview.net/forum?id=HyPeYU9JR6} {Sentence{KV}: Efficient {LLM} inference via sentence-level semantic {KV} caching}.
\newblock In \emph{Second Conference on Language Modeling}.

\end{thebibliography}
\clearpage
\appendix


\begin{CJK*}{UTF8}{gbsn}

\begin{table*}[htbp]
    \centering
    \small
    \renewcommand{\arraystretch}{1.5}
    
    \begin{tabular}{@{} l l p{8cm} @{}}
        \toprule
        \textbf{Priority Level} & \textbf{Category} & \textbf{Details \& Examples} \\
        \midrule
        \textbf{Level-1} & Structural Separators & 
            Paragraphs (\texttt{\textbackslash n\textbackslash n}), \newline
            Markdown (\texttt{---}, \texttt{***}, \texttt{```}), \newline
            Structural Language (\texttt{\}}, \texttt{]}, \texttt{>}) \\
        \midrule
        \textbf{Level-2} & Grammatical \& Semantic & 
            Sentence Terminators (., ?, !, 。？ ！), \newline
            Single Newline (\texttt{\textbackslash n}) \\
        \midrule
        \textbf{Level-3} & Phrasal Separators & 
            English Punctuation (,, ;, :), \newline
            Chinese Punctuation (， ； ： 、) \\
        \midrule
        \textbf{Level-4} & Whitespace & 
            Spaces, Tabs \\
        \bottomrule
    \end{tabular}
    \caption{Hierarchical classification of text separators for recursive chunking, ordered by semantic priority from structural delimiters to whitespace.}
    \label{tab:separators}
\end{table*}
\end{CJK*}

\section{Implementation Details}
\label{appendix:implement}
All experiments are conducted on NVIDIA H20 GPUs. We implement a series of high-performance kernels based on CUDA C++ to maximize hardware utilization and memory access efficiency. Building upon ClusterKV~\cite{clusterkv}, we specifically redesigned the logic for index construction and token selection (retrieval). We further design and write custom kernels for variable-length chunk parallel pooling, label assignment, centroid updates, and radius calculations. Regarding the hyperparameters of~\texttt{\model}, we set the default token budget to 1024. For the chunking strategy, we set the minimum and maximum chunk length thresholds to 8 and 16 tokens, respectively, with a buffer size of 128 tokens. This range is derived from the average information density of natural language sentences and code statements, generally covering a complete semantic unit across most LLMs tokenizers. Unlike rigid fixed-size chunking methods that mechanically cut data and potentially split critical entities (e.g., variable names), our method ensures that atomic components remain intact. Moreover, the algorithm actively searches for the best syntactic breakpoints (e.g., commas or operators) within the effective window to ensure syntactically safe segmentation (Appendix~\S\ref{app:nd}). For the clustering mechanism, we employ spherical $k$-means using the inner product as the distance metric, with the number of iterations fixed at 10.  We observe that the initialization and the number of convergence iterations have a negligible impact on the final performance. Specifically, setting the iteration count to 10-20 yields similar results. To determine the initial number of cluster centers, we set the average fine cluster size to 2 chunks, while limiting the maximum number of coarse units to 64. To ensure stability and align with settings in previous work, we retain full key-value pairs for the first 2 layers and set the attention sink size to 16. All baselines were reproduced based on their official open-source code and recommended parameters.

\section{Details of Natural Delimiters}
\label{app:nd}
We define a hierarchical set of separators to guide the recursive splitting process, as detailed in Table~\ref{tab:separators}. The separators are categorized into four priority levels based on their semantic strength. Level 1 identifies major structural boundaries, such as paragraph breaks and syntax-specific delimiters found in Markdown or code. Level 2 preserves sentence integrity through terminal punctuation and single newlines. Level 3 addresses intra-sentence pauses using phrasal punctuation, while Level 4 serves as a fine-grained fallback based on whitespace. This prioritization ensures that the text is always split at the most logically significant points available before resorting to finer granularity. 
While our method utilizes natural language separators, it is designed to be domain-agnostic and robust. First, our hierarchy explicitly prioritizes structural markers common in technical data, such as Markdown code fences and JSON delimiters (Level 1). Second, the chunking algorithm employs a greedy accumulation strategy with a minimum size constraint, ensuring that code snippets with frequent punctuation are not over-segmented but grouped into coherent logical units (e.g., function bodies). Finally, for unstructured or minified inputs where separators are absent, the method degrades to fixed-size chunking, ensuring it performs no worse than traditional baselines (e.g., Quest~\cite{quest}) even in adversarial scenarios.
Additionally, while SentenceKV~\cite{zhu2025sentencekv} focuses on sentence-level retrieval, its reliance on explicit punctuation limits its applicability to structured data (e.g., code, JSON), which is a core capability of our domain-agnostic \texttt{\model}. Therefore, we prioritize comparisons with general-purpose methods like Quest and ClusterKV.

\begin{figure}[t!]
    \centering
    \includegraphics[width=\linewidth]{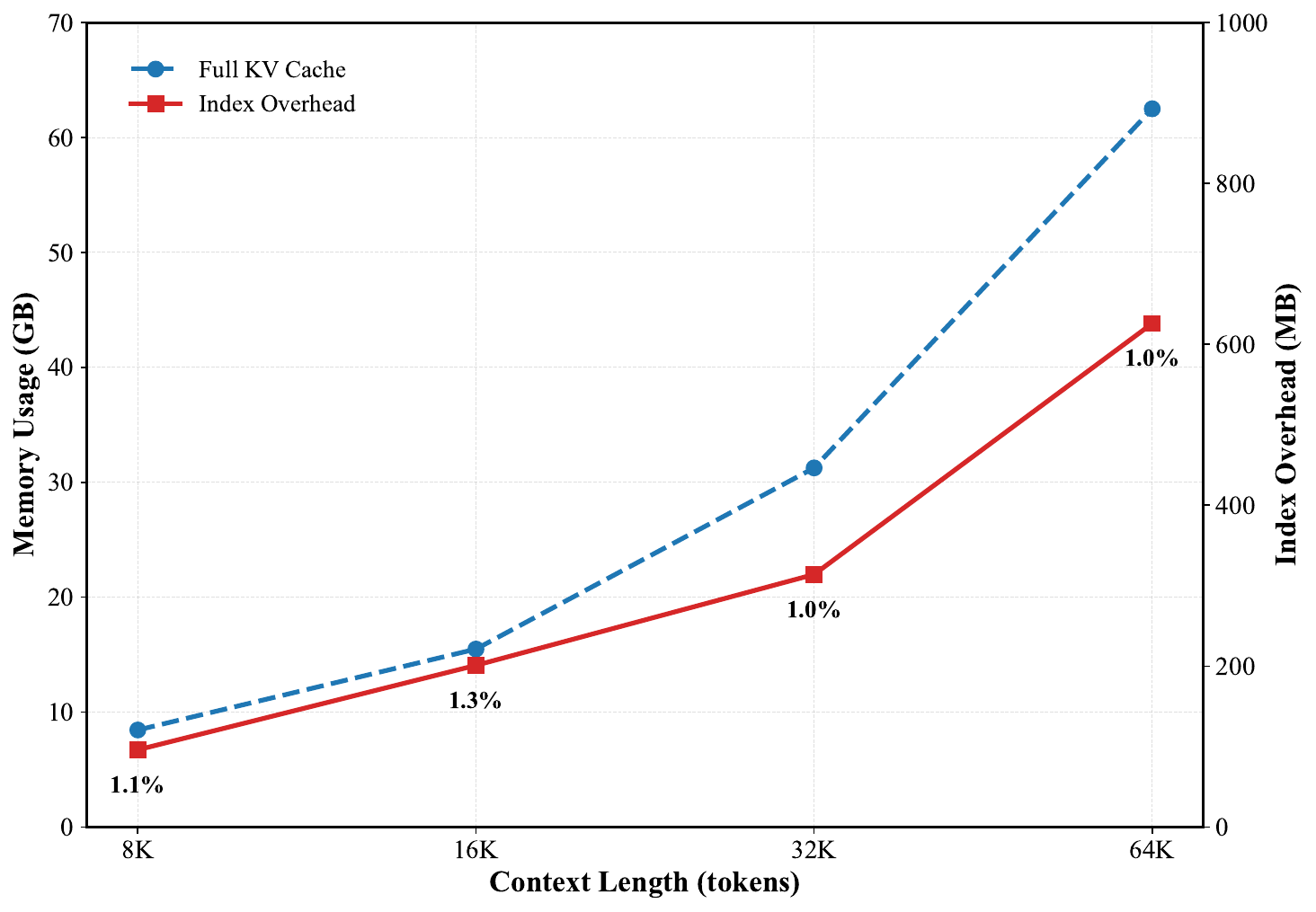}
    \caption{Memory usage comparison between the full KV cache (left axis, GB) and  index overhead of \texttt{\model} (right axis, MB). The percentages annotate the relative size of the index compared to the full KV cache, showing a consistent $\sim$1\% overhead.}
    \label{fig:memory}
    \vspace{-4mm}
\end{figure}

\section{Memory Overhead Analysis of the Indexing Structure}
\label{sec:memory_overhead}

A critical concern regarding indexing-based sparse attention is whether the auxiliary storage required for the index structure itself becomes a bottleneck in ultra-long context scenarios (e.g., 1M tokens). If the index grows disproportionately, it could potentially negate the memory savings achieved by KV cache compression, particularly on memory-constrained edge devices.
To address this, we evaluated the memory footprint of the \texttt{\model} index compared to the full KV cache on the Llama3.1-8B-Instruct model across increasing context lengths. As shown in Figure~\ref{fig:memory}, the index overhead is negligible.

\begin{itemize}
    \item \textbf{Minimal Storage Footprint:} While the Full KV Cache (blue line, left axis) rapidly consumes gigabytes of memory as context length increases, the index overhead (red line, right axis) remains in the megabyte range.
    \item \textbf{Consistent $\sim$1\% Ratio:} Crucially, the ratio of index size to full KV cache size remains stable at approximately 1\% (ranging from 1.0\% to 1.3\%) across all tested lengths up to 64K.
\end{itemize}

This linear scaling behavior implies that the index structure remains lightweight relative to the massive KV tensors. Consequently, the memory cost of the index is orders of magnitude smaller than the memory reclaimed through sparsification, ensuring that our method remains highly efficient even on hardware with strict memory limitations.

\section{Stability Analysis in Ultra-Long Generation}
\label{sec:stability}

To address concerns regarding index quality decay during extended generation scenarios, we conducted a dedicated stability analysis. We define $S_t$ as the set of active clusters (or indices) retrieved at decoding step $t$. To quantify the consistency of the retrieval process over time, we introduce two stability metrics:

\begin{itemize}
    \item \textbf{Jaccard Similarity:} This metric measures the immediate temporal consistency by calculating the overlap between cluster sets at adjacent steps. A higher value indicates a stable focus without abrupt context shifting.
    \begin{equation}
        J(S_t, S_{t-1}) = \frac{|S_t \cap S_{t-1}|}{|S_t \cup S_{t-1}|}
    \end{equation}
    
    \item \textbf{Window Hit Rate:} This metric assesses the locality of the retrieval by calculating the proportion of currently retrieved clusters that have appeared within a recent history window $w$ (where $w=32$). High values imply that the model consistently attends to a relevant working set of information.
    \begin{equation}
        \text{WindowHit}(S_t) = \frac{|S_t \cap (\bigcup_{i=1}^{w} S_{t-i})|}{|S_t|}
    \end{equation}
\end{itemize}

\begin{figure*}[t]
    \centering
    \includegraphics[width=0.9\linewidth]{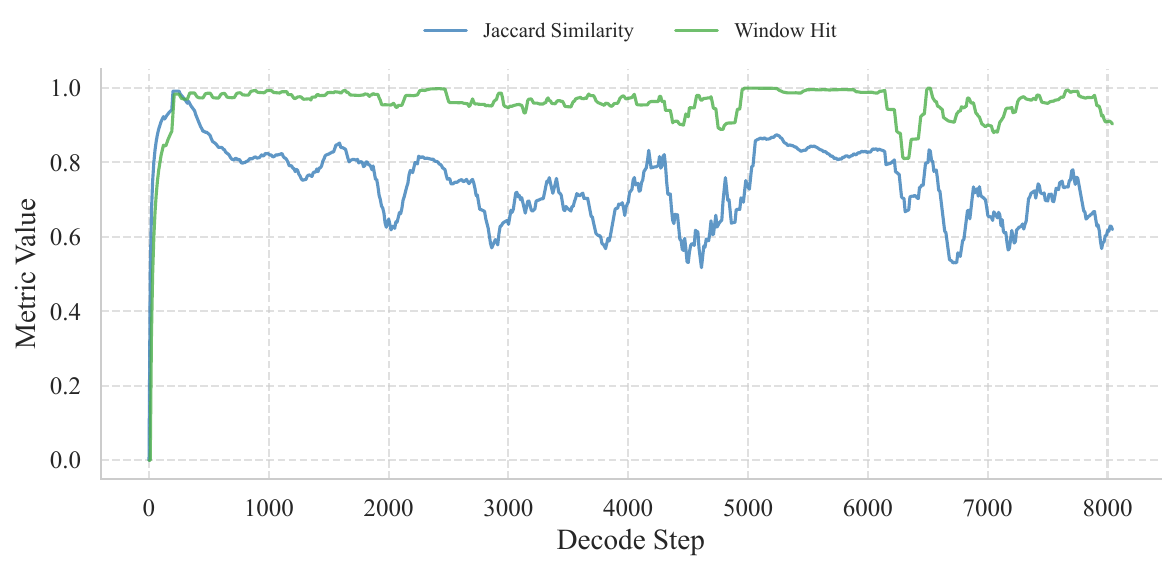}
    \caption{Stability analysis during long-context generation. The blue represents the step-to-step Jaccard Similarity, and the green represents the Window Hit rate ($w=32$). Stability remains robust initially but shows signs of decay after 6k decode steps.}
    \label{fig:stability}
\end{figure*}

As shown in Figure~\ref{fig:stability}, our method demonstrates strong robustness throughout the generation process.
First, the window hit rate (green line) consistently remains close to 1.0 throughout the generation trajectory. This indicates that the model maintains a highly coherent working memory, effectively focusing on relevant historical context without experiencing context loss or generating irrelevant content, even with extremely long sequence lengths.
Second, the Jaccard similarity (blue line) maintains a high baseline level. Although there are some fluctuations, especially after 6k-8k steps, this does not represent model failure. On the contrary, it reflects the  dynamic adaptability of chain-of-thought generation: as the generation topic continuously evolves in long sequences, the retrieved clusters naturally change to adapt to the new semantic content. Importantly, \texttt{\model} avoids catastrophic collapse, maintaining effective retrieval capabilities even in the later stages of generation where previous methods typically experience degradation. Considering that drift of cluster centroids, future work could also perform global re-clustering when necessary~\cite{sun2025rectified}.

\begin{figure}[t!]
    \centering
    \includegraphics[width=\linewidth]{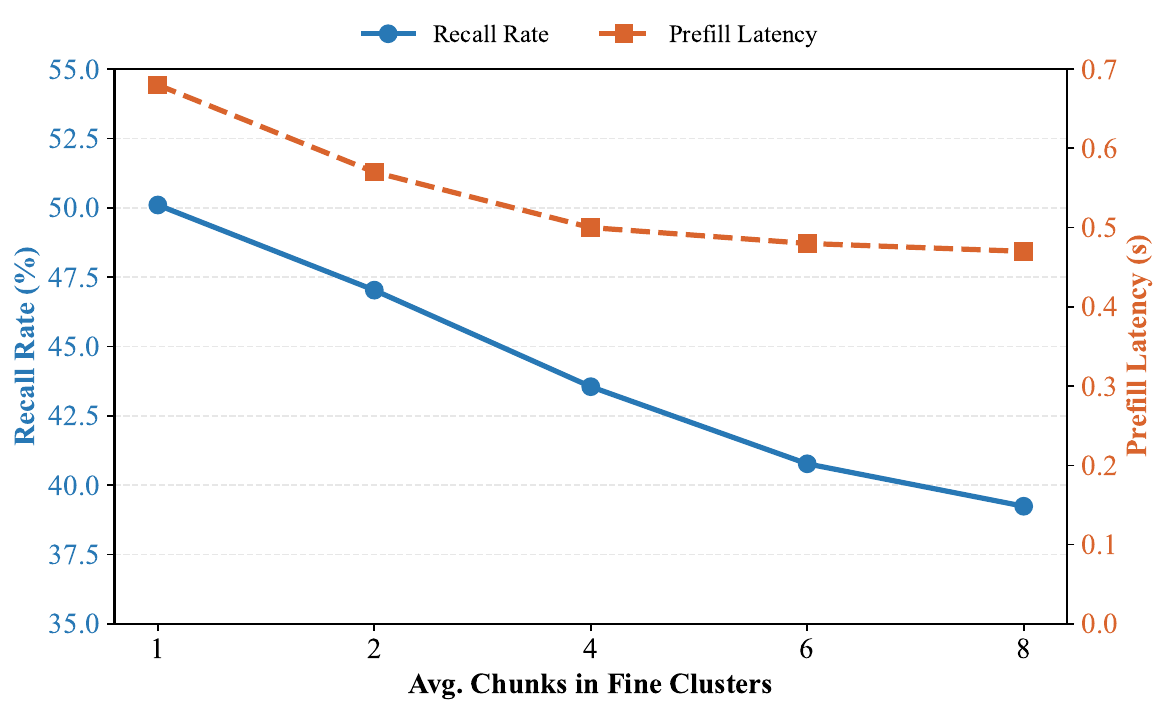}
    \caption{Sensitivity analysis on clustering granularity. We investigate the impact of the average number of chunks per fine cluster on retrieval Recall Rate (blue, left axis) and Prefill Latency (orange, right axis) using the \texttt{gov\_report} subset. }
    \label{fig:sensitivity_clusters}
\end{figure}






\section{Sensitivity Analysis on Clustering Granularity}
\label{sec:sensitivity}

A critical hyperparameter in \texttt{\model} is the granularity of the hierarchical index, specifically the number of fine-grained clusters $L$. Since $L$ naturally scales with the context length, we analyze the impact of the \textit{average cluster size} (i.e., the average number of chunks assigned to each fine cluster) on both retrieval performance and system efficiency. We conducted this sensitivity analysis on the \texttt{gov\_report} subset of LongBench V2. Figure~\ref{fig:sensitivity_clusters} illustrates the trade-off between the recall rate and prefill latency as the average cluster size increases from 1 to 8. The results highlight a clear tension between semantic precision and construction cost:
(1) When the average cluster size is small (e.g., 1 chunk per cluster), the centroid provides a highly accurate representation of the underlying data, resulting in the highest recall ($\sim$50\%). However, as we aggregate more chunks into a single cluster (increasing the x-axis), the centroid becomes a coarser approximation of its constituents. This increased variance within clusters leads to a monotonic decrease in recall, dropping below 40\% when averaging 8 chunks per cluster.
(2) Conversely, the latency of the prefill phase—dominated by the $k$-means clustering overhead—is inversely proportional to the cluster size. Smaller cluster sizes imply a larger number of centroids ($L$) to compute and update, resulting in higher latency ($\sim$0.68s at size 1). Increasing the chunks per cluster reduces $L$, significantly accelerating the index construction process.

Based on these observations, an average of 2 chunks per cluster seems to the optimal setting for our experiments. This configuration strikes a balance, achieving a significant reduction in prefill latency compared to the baseline (size 1) while sacrificing minimal retrieval accuracy ($\sim$3\% drop in recall), whereas larger cluster sizes result in unacceptable performance degradation.

\section{Extended Analysis and Discussions}
\label{sec:extended_analysis}

In this section, we address specific behaviors of \texttt{\model} regarding budget sensitivity and provide a theoretical discussion on the time complexity of our hierarchical retrieval.

\subsection{Behavior in Budget-Sufficient Scenarios}
\label{sec:budget_behavior}
We analyze the behavior of \texttt{\model} when the context length does not strictly exceed the retrieval budget, particularly in reasoning-intensive tasks. A critical design principle of \texttt{\model} is its adaptive nature: the sparse retrieval mechanism is only triggered when the total sequence length (input context + generated chain-of-thought) exceeds the pre-defined token budget.
In scenarios where the sum of the input prompt and the generated chain-of-thought remains within the budget, \texttt{\model} (and ClusterKV) naturally degenerates to full attention. No tokens are pruned, and no approximation error is introduced. This explains the exclusion of ClusterKV in Table~\ref{tab:math500} comparisons. Since the average context length in MATH-500 is relatively short, simple paging or clustering methods often operate in a regime identical to full attention unless an extremely small budget is forced. Consequently, we assert that in short-context tasks without budget constraints, \texttt{\model} incurs zero degradation compared to full attention. It does not suffer from the unnecessary precision loss that affects methods which enforce static sparsity patterns regardless of buffer availability.
However, once the generation extends beyond budget, \texttt{\model} seamlessly transitions to hierarchical retrieval, preserving the critical semantic chunks while strictly adhering to the memory bound.

\subsection{Complexity Analysis: Sub-linear vs. Logarithmic}
\label{sec:complexity_analysis}
A key question regarding hierarchical indexing is the theoretical bound of the retrieval complexity. 

Currently, \texttt{\model} employs a fixed three-tier structure (coarse unit $\to$ fine cluster $\to$ chunk). This design yields a sub-linear time complexity, approximately $O(\sqrt{N})$ depending on the branching factor configuration, rather than strict logarithmic complexity $O(\log N)$.

\paragraph{Why not dynamic depth for $O(\log N)$?} 
While it is theoretically feasible to implement a dynamic tree where the height grows with context length (e.g., a B-tree or hierarchical $k$-means with dynamic depth) to achieve $O(\log N)$ retrieval, we chose a fixed-depth approach for practical efficiency:
\begin{itemize}
    \item \textbf{Overhead Trade-off:} For context windows currently relevant to LLMs (e.g., up to 1M tokens), the constant factors and pointer-chasing overheads introduced by maintaining and traversing a deep, dynamic tree outweigh the theoretical gains of logarithmic complexity.
    \item \textbf{Parallelism:} A flatter, fixed-depth hierarchy maps more efficiently to GPU parallelization (e.g., computing distances to all coarse centroids in parallel) compared to the sequential dependency inherent in traversing a deep tree.
\end{itemize}
Therefore, our current design represents an engineering optimum prioritizing wall-clock speedup over asymptotic optimality.



\begin{figure}[t!]
    \centering
    \includegraphics[width=\linewidth]{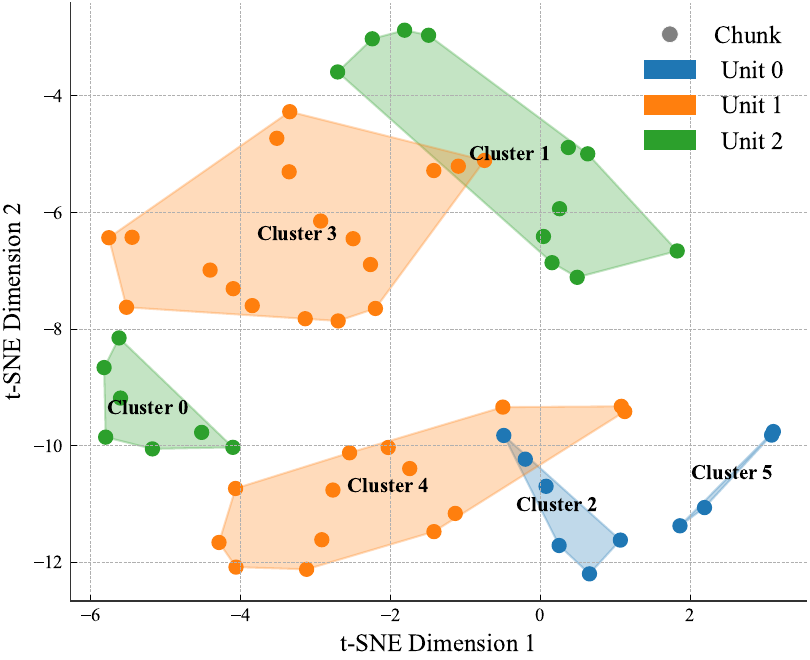}
    \caption{Each point represents a chunk. Colors denote coarse units, while shaded polygons indicate fine clusters.}
    \label{fig:visual_cluster}
\end{figure}
\subsection{Visualization of Hierarchical KV Indexing}
\label{sec:visualization}

To intuitively understand how \texttt{\model} organizes the KV cache in the semantic space, we employ t-Distributed Stochastic Neighbor Embedding (t-SNE) to visualize the hierarchical structure of the indexed chunks. We randomly sampled a sequence from the \textit{Long In-context Learning} task of LongBench V2, extracting the representative key vectors $\bar{k}_j$ of all chunks constructed during the prefill phase.

As illustrated in Figure~\ref{fig:visual_cluster}, we project these high-dimensional vectors into a 2D space. In this visualization:
\begin{itemize}
    \item Each \textbf{point} represents an atomic chunk ($s_j$).
    \item The \textbf{shaded polygons} encompass chunks belonging to the same \textit{fine cluster} ($c_j$).
    \item The \textbf{colors} distinguish different \textit{Coarse Units} ($g_p$, e.g., Unit 0 in blue, Unit 1 in orange).
\end{itemize}

The visualization reveals a clear and well-structured semantic topology. First, chunks within the same fine cluster (e.g., Cluster 5 in Unit 0) are tightly grouped, validating our assumption that structure-aware chunks preserve local semantic coherence. Second, the fine clusters themselves are spatially aggregated into distinct coarse units. For instance, the orange region (Unit 1) effectively covers a broad semantic area containing Clusters 3 and 4, while being clearly separated from the green region (Unit 2).

This strong spatial locality confirms the effectiveness of our pruning strategy: a query vector falling near the orange centroid can safely discard the blue and green branches, as their constituent chunks lie far outside the relevant semantic radius. This hierarchical separation is the cornerstone of \texttt{\model}'s ability to maintain high recall with sub-linear retrieval complexity.

\section{Dataset}
\subsection{LongBench V2}
\begin{figure}[t]
    \centering
    \includegraphics[width=\linewidth]{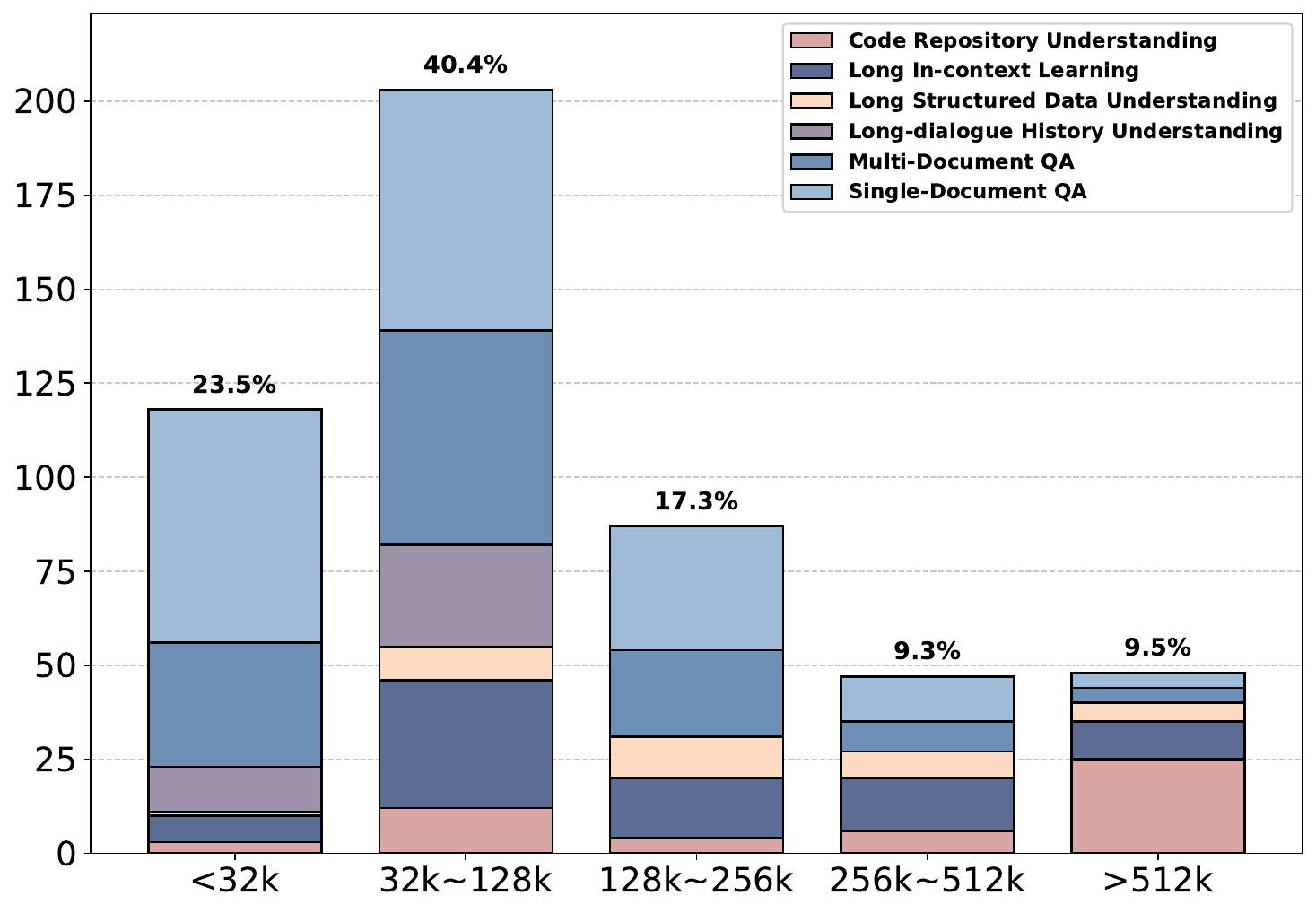}
    \caption{Statistical distribution of data length ranges in the \textsc{LongBench V2} dataset, using the Llama3.1-8B-instruct tokenizer.}
    \label{fig:longbench_dist}
\end{figure}
\textsc{LongBench V2}~\cite{bai2025longbench} is a benchmark designed to evaluate the deep understanding and reasoning capabilities of LLMs on long contexts in real-world multi-task scenarios. The dataset consists of 503 highly challenging multiple-choice questions with context lengths ranging from 8k to 2M words (most are below 128k), covering six major task categories: single-document question answering, multi-document question answering, long-context learning, long dialogue history understanding, code repository understanding, and long structured data understanding. To ensure the breadth and practicality of the data, the questions were constructed by nearly a hundred highly educated individuals with diverse professional backgrounds and underwent rigorous review. The difficulty is extremely high; even human experts using search tools only achieved a 53.7\% accuracy rate within 15 minutes. 
Tables~\ref{tab:expert_performance} show the specific statistical information for the six task categories, and Figures~\ref{fig:longbench_dist} shows the distribution of the dataset across different length ranges. As shown in the figures, \textsc{LongBench V2} is extremely challenging, with its core evaluation range concentrated in the medium-to-long text range of 32k to 128k (40.4\%), while also retaining a considerable proportion of ultra-long text (>512k) test cases, with a maximum text length of up to 4M.

To provide a comprehensive assessment of model performance under varying memory and computational constraints, \textsc{LongBench V2} categorizes its test instances into three distinct length groups based on the input context size: \textit{Short} (0--32k tokens), \textit{Medium} (32k--128k tokens), and \textit{Long} (exceeding 128k tokens, up to 2M). The \textit{Overall} score represents the macro-average across all instances. This stratified evaluation is crucial for identifying the performance degradation of LLMs, as many models exhibit significant degradation when the context exceeds their effective attention window or fine-tuned limits. By isolating these categories, we can distinguish between the basic reasoning capabilities and its robustness in handling extreme-scale information retrieval and synthesis.

\begin{table*}[htbp]
    \centering
    \small
    \begin{tabular}{lccccc}
        \toprule
        Task & \# Samples & Avg. Length & Max Length & Expert Acc. & Expert Time \\
        \midrule
        Code Repository Understanding       & 50  & 1M   & 4M   & 44\% & 6.4 min \\
        Long In-context Learning            & 81  & 259k & 1.5M & 63\% & 8.3 min \\
        Long Structured Data Understanding  & 33  & 390k & 3M   & 73\% & 6.4 min \\
        Long-dialogue History Understanding & 39  & 74k  & 120k & 79\% & 8.2 min \\
        Multi-Document QA                   & 125 & 127k & 1.7M & 36\% & 6.1 min \\
        Single-Document QA                  & 175 & 110k & 860k & 55\% & 8.9 min \\
        \bottomrule
    \end{tabular}
     \caption{Data statistics and expert performance across tasks of \textsc{LongBench V2}.}
    \label{tab:expert_performance}
\end{table*}

\begin{table*}[h!]
\centering

\small
\begin{tabular}{cccccccccc}
\toprule
Context / Task & single & multikey & multivalue & multiquery & vt & fwe & qa1 & qa2 & Avg. \\
\midrule
\multicolumn{10}{c}{Full Attention} \\
\midrule
4k  & 100.0 & 100.0 & 100.0 & 100.0 & 82.5 & 84.4 & 89.8 & 59.2 & 89.5 \\
8k  & 91.8  & 100.0 & 100.0 & 99.0  & 79.6 & 87.1 & 81.6 & 59.2 & 87.3 \\
16k & 63.3  & 100.0 & 98.5  & 98.5  & 82.0 & 93.9 & 79.6 & 51.0 & 83.3 \\
32k & 95.9  & 95.9  & 97.5  & 97.5  & 76.3 & 76.9 & 79.6 & 59.2 & 84.8 \\
\midrule
\multicolumn{10}{c}{LycheeCluster} \\
\midrule
4k  & 100.0 & 100.0 & 100.0 & 99.5  & 81.2 & 80.3 & 89.8 & 59.2 & 88.8 \\
8k  & 100.0 & 100.0 & 96.4  & 100.0 & 78.0 & 83.7 & 81.6 & 59.2 & \textbf{87.4} \\
16k & 93.9  & 100.0 & 94.9  & 94.9  & 73.1 & 89.8 & 77.6 & 51.0 & \textbf{84.4} \\
32k & 95.9  & 98.0  & 92.9  & 93.9  & 85.7 & 76.2 & 75.5 & 59.2 & 84.7 \\
\bottomrule
\end{tabular}%
\caption{Performance comparison on RULER benchmark. Note that LycheeCluster achieves comparable or even superior average performance compared to Full Attention, especially at 16k length.}
\label{tab:ruler_results}
\end{table*}
\subsection{MATH500} 
To evaluate the advanced mathematical reasoning capabilities of LLMs, we utilize the \textsc{MATH500} benchmark~\cite{lightman2023let}. This dataset is a high-quality subset of the original MATH benchmark \cite{hendrycks2021measuring}, consisting of 500 problems curated to represent a broad spectrum of difficulty levels and mathematical topics. The problems are sourced from elite mathematics competitions, including the American Mathematics Competitions (AMC 10/12) and the American Invitational Mathematics Examination (AIME). MATH-500 spans seven core subjects: Algebra, Geometry, Number Theory, Counting \& Probability, Prealgebra, Intermediate Algebra, and Precalculus. Due to its requirement for multi-step logical deduction and precise symbolic manipulation, MATH-500 has become a standard metric for assessing the chain-of-thought reasoning performance of state-of-the-art models.

\section{More Experiment Results on RULER Benchmark}
\label{app:ruler}
To assess the long-context capabilities, we evaluate our method on the RULER benchmark \cite{hsieh2024ruler}. RULER serves as a comprehensive testbed for long-context language models, expanding beyond simple retrieval to include complex aggregation and reasoning tasks. We report results on a diverse set of tasks: \textit{single}, \textit{multikey}, \textit{multivalue}, \textit{multiquery}, \textit{vt}, \textit{fwe}, \textit{qa1}, and \textit{qa2}. We compare \textbf{LycheeCluster} against the \textbf{Full Attention} baseline across sequence lengths from 4k to 32k.

The experimental results are presented in Table~\ref{tab:ruler_results}. Our method demonstrates remarkable stability and robustness across all context lengths. In the 4k setting, LycheeCluster achieves an average score of 88.8, comparable to the Full Attention baseline. Notably, as the sequence length extends, LycheeCluster exhibits superior stability in specific scenarios. At 16k length, our method outperforms the full attention model, primarily because the full attention baseline suffers a significant performance drop in the \textit{single} task (dropping to 63.3), whereas LycheeCluster maintains a high score of 93.9. Even at the extreme length of 32k, our approach maintains parity with the full attention mechanism, proving that LycheeCluster can effectively handle extensive contexts without compromising performance.

\end{document}